\pdfoutput=1

\documentclass[11pt]{article}

\usepackage[review]{acl}

\usepackage{times}
\usepackage{latexsym}

\usepackage[T1]{fontenc}

\usepackage[utf8]{inputenc}

\usepackage{microtype}

\usepackage{inconsolata}

\usepackage{amsmath,amsfonts,bm}

\def\eqref#1{equation~\ref{#1}}

\def\1{\bm{1}}

\DeclareMathAlphabet{\mathsfit}{\encodingdefault}{\sfdefault}{m}{sl}
\SetMathAlphabet{\mathsfit}{bold}{\encodingdefault}{\sfdefault}{bx}{n}

\usepackage{hyperref}
\usepackage{url}

\usepackage{graphicx}
\usepackage{hyperref}       
\usepackage{url}            
\usepackage{booktabs}       
\usepackage{amsfonts}       
\usepackage{nicefrac}       
\usepackage{microtype}      
\usepackage{xcolor}         
\usepackage{amssymb}
\usepackage{amsmath}
\usepackage{multirow}
\usepackage{makecell}
\usepackage{wrapfig}
\usepackage{subcaption}
\usepackage{bm}
\usepackage{bbding}
\usepackage{colortbl}
\usepackage{bbding}
\newcommand{\modelname}{EasyGen}

\definecolor{mygray}{gray}{.9}
\newcommand{\tune}{\includegraphics[width=.25cm]{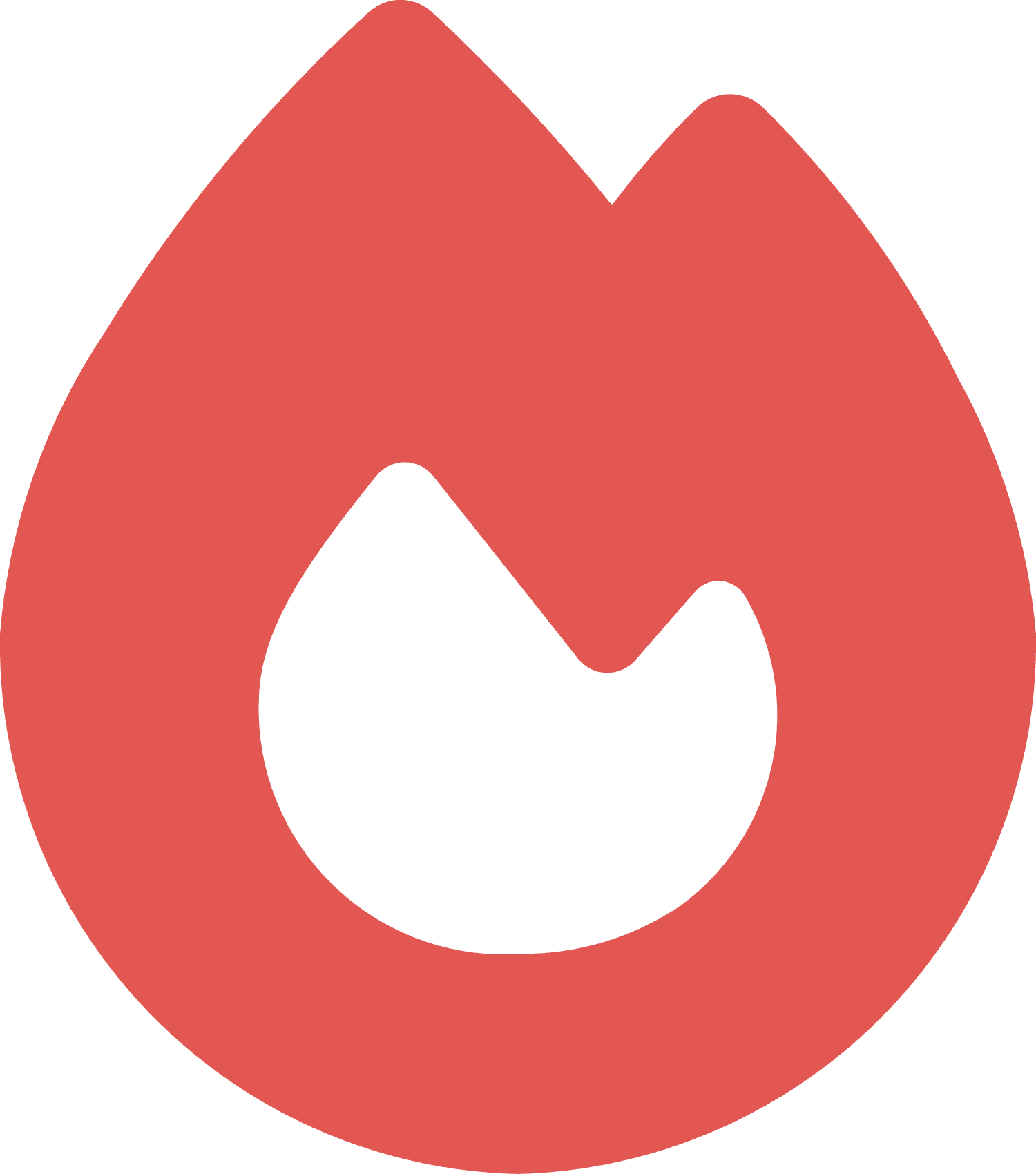}}
\newcommand{\freeze}{\includegraphics[width=.25cm]{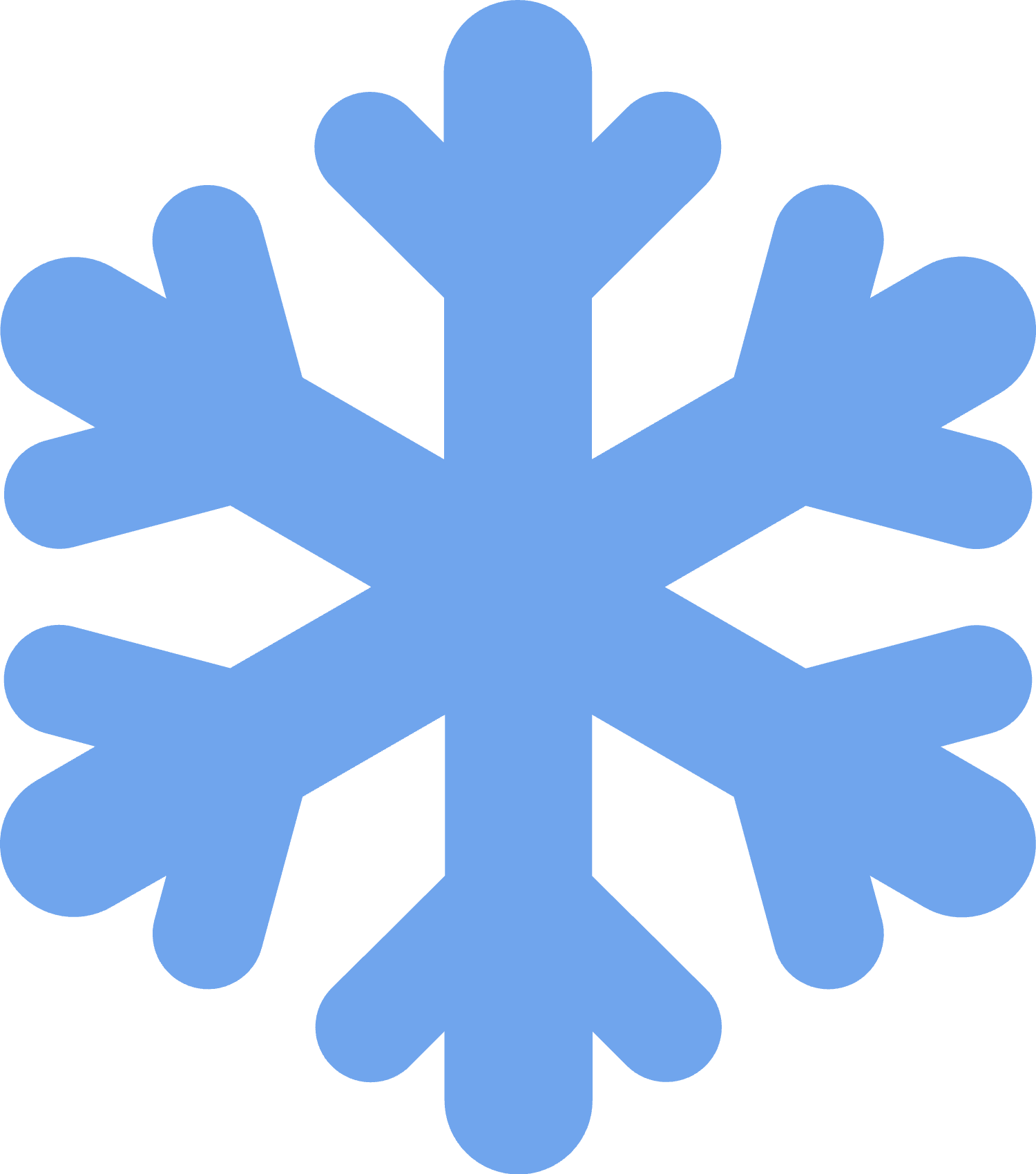}}

\title{
EasyGen: Easing Multimodal Generation with BiDiffuser and LLMs
}

\author{ Xiangyu Zhao, Bo Liu\footnotemark[1], Qijiong Liu\footnotemark[1], Guangyuan Shi\footnotemark[1], Xiao-Ming Wu\textsuperscript{\Envelope} \\Department of Computing, The Hong Kong Polytechnic University\\ \{xiang-yu.zhao, bokelvin.liu, jyonn.liu, guang-yuan.shi\}@connect.polyu.hk, \\xiao-ming.wu@polyu.edu.hk}

\begin{document}

\maketitle

\renewcommand{\thefootnote}{\fnsymbol{footnote}}
\footnotetext[1]{Co-Second Author.}


\begin{abstract}

We present EasyGen, an efficient model designed to enhance multimodal understanding and generation by harnessing the capabilities of diffusion models and large language models (LLMs). Unlike existing multimodal models that predominately depend on encoders like CLIP or ImageBind and need ample amounts of training data to bridge modalities, EasyGen leverages BiDiffuser, a bidirectional conditional diffusion model, to foster more efficient modality interactions. \modelname{} achieves text generation by training a projection layer linking BiDiffuser and an LLM, and facilities image generation by training an adapter to align the LLM’s text space with the BiDiffuser’s image space. Comprehensive quantitative and qualitative experiments show that \modelname{} excels in data-efficient training, high-quality image generation, and extendibility, effectively addressing the challenges in multimodal generation. The source code is available at \url{https://github.com/zxy556677/EasyGen}.

\end{abstract}

\section{Introduction}

In recent years, remarkable progress has been made in the field of artificial intelligence generated content (AIGC), notably in technologies like large language models (LLMs)~\citep{vicuna2023,touvron2023llama,brown2020language,chowdhery2022palm,zeng2022glm} for text generation and diffusion models~\cite{rombach2022high, nichol2022glide,saharia2022photorealistic} for visual generation. 
These breakthroughs have paved the way for the development of multimodal large language models (MLLMs), sparking a recent trend of incorporating extra visual modules into LLMs.
Collaborative models, such as Visual ChatGPT~\citep{wu2023visual} and MM-REACT~\citep{yang2023mm}, strategically use externally pre-trained tools to translate visual information into text descriptions and feed the data into LLMs.
However, they are exclusively dependent on pre-trained tools for inference.
Contrarily, end-to-end trained models including the BLIP series~\citep{li2023blip}, LLaVA series~\citep{liu2023visual,liu2023improved}, MiniGPT-4~\citep{zhu2023minigpt}, and mPLUG-Owl~\citep{ye2023mplug} focus on mapping image information to the text space of LLMs, enabling LLMs to comprehend visual inputs. 


Existing end-to-end models are also not without limitations. First, most of these multimodal models rely on either CLIP~\citep{radford2021learning} or ImageBind~\citep{girdhar2023imagebind} as their image encoder. While these encoders excel in learning unified representations that encompass both text and images, they face challenges when it comes to transforming between different modalities. This predicament makes current vision-language models relying heavily on sizable data sets to align CLIP/Bind-encoded images with the language model, due to the disparity between different modalities.

Moreover, the majority of previous multimodal models have concentrated on comprehending multimodal content and lacked the capability to generate multimodal responses, such as content beyond text. 
Several concurrent works, such as Emu (Sun et al., 2023) and NExT-GPT (Wu et al., 2023), have utilized diffusion models for multimodal generation. Typically, these methods involve training a projection layer to align the output embedding space of the LLM with the input embedding space of the diffusion model (encoded by CLIP’s text encoder) using an MSE loss. However, this approach may lead to the underutilization of the semantic understanding and reasoning capabilities of the LLM, and may introduce information loss in the alignment process, ultimately leading to lower image generation quality compared to the original diffusion model, as elaborated in Sec.~\ref{sec:t2i} and Tab.~\ref{tab:fid distance}.

\begin{figure*}[t]
	\centering
	\includegraphics[width=0.9\textwidth]{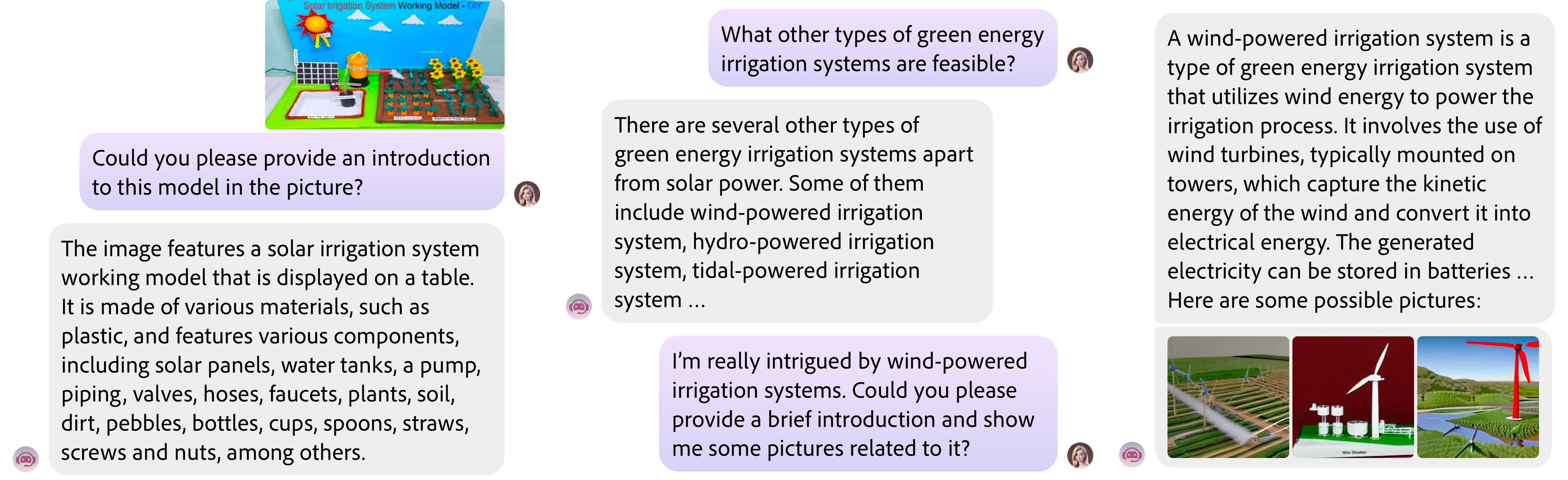}
        \caption{Our model EasyGen can understand multimodal inputs and generate multimodal responses, as illustrated by model-generated speech bubbles in grey color, which include both text and images.
        }
        \label{introduction_example}
\end{figure*}



\begin{figure}[t]
	\centering
	\includegraphics[width=.48\textwidth]{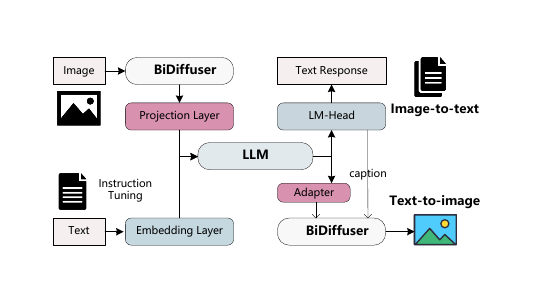}
        \caption{Overview of EasyGen.
        }
        \label{overview}
\end{figure}

In this work, we propose \modelname{}, an end-to-end model that facilitates multimodal generation with a single bidirectional conditional diffusion model and LLMs, as illustrated in Figure~\ref{overview}. The diffusion model, called BiDiffuser, is obtained by fine-tuning the UniDiffuser~\citep{bao2023one}, with a specific focus on targeted image-to-text and text-to-image tasks. This fine-tuning addresses UniDiffuser's limitation of attempting to fit all conditional distributions, including those based on noisy inputs, into a single model, which reduces its effectiveness on specific tasks like conditional generation from noise-free inputs. BiDiffuser plays a pivotal role for both text and image generation. In \modelname{}, text generation is achieved by training a projection layer that connects BiDiffuser and an LLM, while image generation is facilitated by training an adapter that infuses the text representation of the LLM into BiDiffuser. Figure~\ref{introduction_example} showcases \modelname{}'s ability to handle multimodal inputs and generate appropriate multimodal responses. 

\modelname{} holds three significant advantages that address the challenges in multimodal generation:

    First, \modelname{} offers competitive performance in a \emph{data-efficient} way compared to cutting-edge models, as shown in \textbf{Tab.~\ref{tab:representation distance} (Sec.~\ref{data_efficiency})}. This is due to BiDiffuser's ability to simplify the alignment of its embedding space with an LLM, which allows for efficient training with less data for image-to-text tasks such as image captioning and VQA.
    
    Second, \modelname{} exhibits \emph{superior image generation quality}, surpassing other end-to-end MLLMs, as shown in \textbf{Tab.~\ref{tab:fid distance} (Sec.~\ref{sec:t2i})}. This is attributed to the adapter's design (Sec.~\ref{sec: sur_adapter}), which aligns the LLM's text space with the diffusion model's image space, thereby utilizing the LLM's semantic understanding and reasoning capabilities. In contrast, the projection layers in other MLLMs like NExT-GPT only align the LLM's text space with the diffusion model's text space and are not trained by the image denoising objective.

    Third, \modelname{} can be \emph{readily adapted} to manage complex vision-language tasks by incorporating more advanced visual encoders or by integrating BiDiffuser into contemporary sophisticated multimodal LLMs like LLaVA to enhance performance, as shown in \textbf{Tab.~\ref{VQA results} (Sec.~\ref{integrating MLLM})}.

\section{Related Work}

\textbf{Multimodal Language Models}. 
Recent research has witnessed a surge of interest in multimodal LLMs, including collaborative models~\citep{wu2023visual, yang2023mm, shen2023hugginggpt} and end-to-end methods~\citep {alayrac2022flamingo, guo2022images, li2022blip, bao2021beit, wang2022image, wang2022ofa, wang2022ofa}. More recently, some works also explore training LLMs with parameter-efficient tuning~\citep{li2023blip, zhang2023llama} and instruction tuning~\citep{instructblip, liu2023visual, ye2023mplug, zhu2023minigpt, li2023otter}. Different from them, \modelname{} is built upon BiDiffuser, which promotes more efficient interactions between modalities.

\noindent\textbf{Multimodal Diffusion Models}. Diffusion generative models~\citep{rombach2022high,ramesh2021zero,nichol2022glide,ruiz2023dreambooth} have achieved strong results in text conditioned image generation works. Specifically, Versatile Diffusion~\citep{xu2023versatile} employs the U-Net~\citep{ronneberger2015u} architecture with a multi-flow design to tackle multiple modalities and tasks, while UniDiffuser~\citep{bao2023one} adopts the U-ViT~\citep{bao2023all} framework to treat both image and text as sequential token streams for diffusion calculations. However, these models are unable to complete complex language tasks. \modelname{} combines the advantages of diffusion models and LLMs and achieves competitive performance in both image-to-text and text-to-image tasks.

\textbf{Multimodal Response Generation}. Recent research has made significant advancements in multimodal response generation~\citep{koh2023grounding, tang2023any, zhang2023meta, wu2023next, pan2023kosmos, koh2023generating, sun2023generative, dong2023dreamllm} using text-to-image models such as Stable Diffusion. However, the lack of semantic understanding capability in the CLIP text encoder may result in low-quality generated images. \modelname{} addresses this issue by transferring knowledge from LLM to BiDiffuser via an adapter, enabling the creation of high-quality textual semantic representations for text-to-image generation.

\section{Basics of Diffusion Models}



\textbf{Unconditional Generation.} Given a data sample taken from a real data distribution $\mathbf{x}_{0} \sim q(\mathbf{x}_0)$, diffusion models~\citep{sohl2015deep,ho2020denoising} first destruct the data by constructing a Markov forward process and gradually injecting noise to the data:
\begin{equation}
	\begin{aligned}
q(\mathbf{x}_{1:T}|\mathbf{x}_0)&=\prod_{t=1}^{T}q(\mathbf{x}_t|\mathbf{x}_{t-1}), \\ \quad q(\mathbf{x}_{t}|\mathbf{x}_{t-1}) &= \mathcal{N}(\mathbf{x}_{t};\sqrt{1-\beta_{t}}\mathbf{x}_{t-1},\beta_{t}\mathbf{I}), 
    \end{aligned}\label{eq:forward}
\end{equation}
%
where $\beta_{t} \in (0,1)$ is the variance added at diffusion step $t$.
Then, they learn to reverse the process:
\begin{equation}
\begin{aligned}
    p(\mathbf{x}_{0:T}) &= p(\mathbf{x}_T)\prod_{t=1}^{T}p_{\theta}(\mathbf{x}_{t-1}|\mathbf{x}_t), \\ \quad p_{\theta}(\mathbf{x}_{t-1}|\mathbf{x}_{t}) &= 
    \mathcal{N}(\mathbf{x}_{t-1}; \mu_{t}(\mathbf{x}_{t},t),\sigma_t^{2}\mathbf{I}),
\end{aligned}
\end{equation}
where $p(\mathbf{x}_T)=\mathcal{N}(\mathbf{x}_T; 0,\mathbf{I})$ is  
the standard Gaussian distribution and $\mu_{t}(\cdot)$ is the parameterization of the predicted mean. Diffusion models are trained to maximize the marginal likelihood of the data $\mathbb{E} [\log p_{\theta}(\mathbf{x}_{0})]$, and the canonical objective is the variational lower bound of $\log p_{\theta}(\mathbf{x}_{0})$. 
Denoising diffusion probabilistic models~\citep{ho2020denoising} generate samples $\mathbf{x}_t
\sim q(\mathbf{x}_t|\mathbf{x}_0)$ by injecting noise $\bm{\epsilon}\sim \mathcal{N}(0, \mathbf{I})$ to the data $\mathbf{x}_0$, and train a network $\bm{\epsilon}_{\theta}(\cdot)$ to predict the added noise $\bm{\epsilon}$ using a standard mean squared error loss:
\begin{equation}
        \label{unconditional loss}
	\begin{aligned}
		\mathcal{L} := \mathbb{E}_{\mathbf{x}_{0},{\bm{\epsilon}},t}[\Vert {\bm{\epsilon}} - \bm{\epsilon}_{\theta}(\mathbf{x}_{t},t) \Vert^{2}].
	\end{aligned}
\end{equation}

\paragraph{Conditional Generation.} 
For conditional generation, a paired data $(\mathbf{x}_{0}, \mathbf{y}_{0})$ is given, and the aim is to model the conditional data distribution $q(\mathbf{x}_{0}|\mathbf{y}_{0})$, where $\mathbf{y}_{0}$ can be image class or text prompt. Conditional generation includes classifier guidance~\citep{dhariwal2021diffusion} and classifier-free guidance~\citep{ho2021classifier}. 
Classifier guidance requires training an extra classifier on noisy data at inference time to improve sample quality. 
For classifier-free guidance, 
no classifier needs to be trained. The denosing network $\bm{\epsilon}_{\theta}(\mathbf{x}_{t}|\mathbf{y}_{0})$ simply conditions on the information encoded in $\mathbf{y}_0$. At inference time, with a guidance scale $s$, the modified score estimate is further in the direction of $\bm{\epsilon}_{\theta}(\mathbf{x}_{t}|\mathbf{y}_{0})$ and away from the unconditional model $\bm{\epsilon}_{\theta}(\mathbf{x}_{t}|\emptyset)$ ($\emptyset$ is a null token):
%
%
%
\begin{equation}
	\begin{aligned}
		\hat{\bm{\epsilon}}_{\theta}(\mathbf{x}_{t}|\mathbf{y}_{0}) = \bm{\epsilon}_{\theta}(\mathbf{x}_{t}|\emptyset) + s \cdot (\bm{\epsilon}_{\theta}(\mathbf{x}_{t}|\mathbf{y}_{0}) - \bm{\epsilon}_{\theta}(\mathbf{x}_{t}|\emptyset)). \nonumber
	\end{aligned}
\end{equation}


\section{Proposed Model: EasyGen
}


        
    

We propose \modelname{}, a model capable of processing multimodal inputs and generating multimodal outputs. It achieves easy multimodal generation by leveraging a bidirectional conditional diffusion model to effectively bridge the gap between different modalities and an LLM to comprehend multimodal tasks and produce textual responses containing cues for multimodal message creation. In the subsequent section, we outline the multimodal generation process of \modelname{}.  


\subsection{
Pre-training BiDiffuser: A Bidirectional Conditional Diffusion Model
}
\label{diffusion module}

Since the text space of LLMs is discrete, to minimize the disparity between the output of a diffusion model and the input of LLMs, we leverage Unidiffuser, a unified diffusion model capable of transforming images into the discrete text space. During the training process, UniDiffuser injects noise $\bm{\epsilon}^{x}$ and $\bm{\epsilon}^{y}$ to a set of paired image-text data $(\mathbf{x}_0, \mathbf{y}_0)$ and generates noisy data $\mathbf{x}_{t^{x}}$ and $ \mathbf{y}_{t^{y}}$, where $0\leqslant t^{x},t^{y}\leqslant T$ represent two individual timesteps (perturbation levels). It then trains a joint denoising transformer U-ViT~\citep{bao2023all} $\bm{\epsilon}_{\theta}(\mathbf{x}_{t^{x}},\mathbf{y}_{t^{y}},t^{x},t^{y})$ to predict the noise $\bm{\epsilon}^{x}$ and $\bm{\epsilon}^{y}$ by minimizing the mean squared error loss:
\begin{equation}
	\begin{aligned}
		\mathbb{E}_{\bm{\epsilon}^{x},\bm{\epsilon}^{y},\mathbf{x}_{0},\mathbf{y}_{0}}[\Vert [\bm{\epsilon}^{x},\bm{\epsilon}^{y}] - \bm{\epsilon}_{\theta}(\mathbf{x}_{t^{x}},\mathbf{y}_{t^{y}},t^{x},t^{y}) \Vert^{2}],\nonumber
	\end{aligned}\label{eq:Unidiffuser}
\end{equation}
where the output of $\bm{\epsilon}_{\theta}$ is the concatenation of the estimated noise $\bm{\epsilon}_{\theta}^x$ and $\bm{\epsilon}_{\theta}^y$, i.e., $\bm{\epsilon}_{\theta}=[\bm{\epsilon}_{\theta}^x,\bm{\epsilon}_{\theta}^y]$.

By predicting $\bm{\epsilon}_{\theta}(\mathbf{x}_{t^{x}},\mathbf{y}_{t^{y}},t^{x},t^{y})$ for any $t^{x}$ and $t^{y}$, UniDiffuser learns all distributions related to $(\mathbf{x}_0, \mathbf{y}_0)$ simultaneously. This includes all conditional distributions: $q(\mathbf{x}_0|\mathbf{y}_0)$ for text-to-image generation, $q(\mathbf{y}_0|\mathbf{x}_0)$ for image-to-text generation, and those conditioned on noisy input, i.e., $q(\mathbf{x}_0|\mathbf{y}_{t^{y}})$ and $q(\mathbf{y}_0|\mathbf{x}_{t^{x}})$, for $0<t^{x},t^{y}\leq T$.  
Learning a conditional distribution $q(\mathbf{x}_0|\mathbf{y}_{t^{y}})$ or $q(\mathbf{y}_0|\mathbf{x}_{t^{x}})$ can be seen as learning a distinct task. From a multitask learning perspective, due to limited network capacity, learning many tasks simultaneously (i.e., fitting all distributions to a single network) may result in \emph{task competition or task conflict}, ultimately leading to suboptimal performance in particular tasks such as $q(\mathbf{x}_0|\mathbf{y}_0)$ and $q(\mathbf{y}_0|\mathbf{x}_0)$.

\begin{figure}[t]
    \centering
    \includegraphics[width=.99\linewidth]{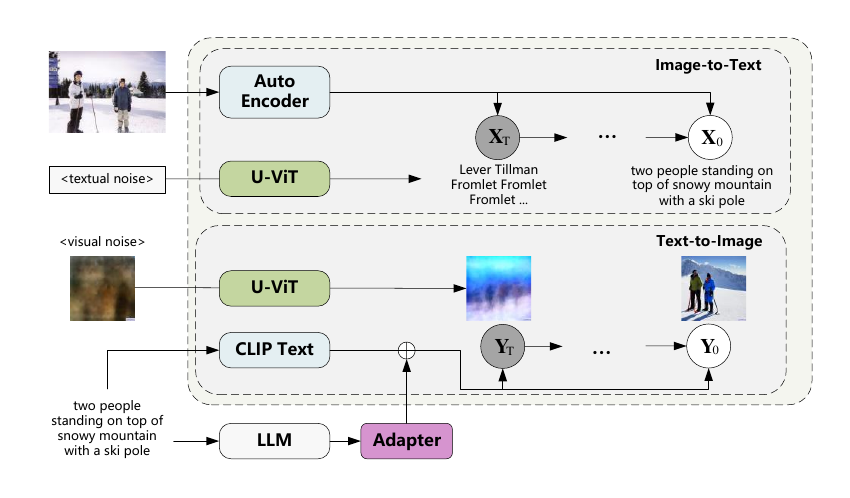}
    \caption{The training of BiDiffuser involves finetuning the denoising transformer U-ViT in UniDiffuser with a joint objective of image-to-text and text-to-image tasks. 
    }
    \label{fig:unidiffuser}
\end{figure}

To resolve this issue and enhance the performance of both image-to-text and text-to-image generation tasks, we finetune UniDiffuser with exclusive emphasis on the two tasks:
\begin{equation}
	\begin{aligned}
		\mathcal{L}_{d} = 
  \mathbb{E}_{\bm{\epsilon}^{x},\bm{\epsilon}^{y},\mathbf{x}_{0},\mathbf{y}_{0}}[\Vert
  \bm{\epsilon}^{x} - \bm{\epsilon}_{\theta}^{x}(\mathbf{x}_{t^{x}},\mathbf{y}_0,t^{x},0)\Vert^{2} 
  + \\ \alpha 
  \Vert \bm{\epsilon}^{y} - \bm{\epsilon}_{\theta}^{y}(\mathbf{x}_0,\mathbf{y}_{t^{y}},0,t^{y})\Vert^{2}].\nonumber
	\end{aligned}
 \label{eq:L_easyGen}
\end{equation}
where $\alpha$ is a hyperparameter to balance the learning paces of the two tasks. 
As depicted in Figure~\ref{fig:unidiffuser}, our training objective entails predicting the text $\mathbf{y}_{0}$ based on the input image $\mathbf{x}_{0}$ and vice versa, where the input conditions for the model are noise-free. We name the finetuned model ``BiDiffuser'', signifying its specialized ability in bidirectional conditional generation.

\subsection{Pre-training an Adapter to Enhance BiDiffuser's SUR Capability }
\label{sec: sur_adapter}
BiDiffuser uses the text encoder of CLIP, which is trained with image-text contrastive learning, limiting its semantic understanding and reasoning (SUR) ability for image generation. Drawing inspiration from \citet{zhong2023adapter}, we utilize LLMs to enhance the SUR capability of BiDiffuser. 
Specifically, we design an adapter that employs the attention mechanism to integrate the semantic information from LLM's last hidden state $ f_\text{LLM}(\cdot)$ into the CLIP text encoder $ f_\text{CLIP}(\cdot)$. 
The adapter consists of a projection layer $\text{MLP}(\cdot)$ and a cross-attention layer $\text{Att}(\cdot)$. Given a paired image-text data  $(\mathbf{x}_0, \mathbf{y}_0)$, we can get $y_\text{sur}$ with enhanced SUR via the adapter:
\begin{equation}
	\begin{aligned}
            y_\text{sur} = \text{Att}(f_{\text{CLIP}}(\mathbf{y}_{0})W^{Q}, \text{MLP}(f_{\text{LLM}}(\mathbf{y}_{0}))W^{K}, \\\text{MLP}(f_{\text{LLM}}(\mathbf{y}_{0}))W^{V}). \nonumber
	\end{aligned}
 \label{eq:adapter}
\end{equation}
Then, the semantic input to BiDiffuser is the combination of $y_\text{sur}$ and the CLIP text encoding of $\mathbf{y}_0$:
\begin{equation}
	\begin{aligned}
            y_{0} = \lambda \cdot y_\text{sur} + (1-\lambda) \cdot f_{\text{CLIP}}(\mathbf{y}_{0}),
	\end{aligned}
 \label{eq:fuse}
\end{equation}
where $\lambda$ is a balancing parameter. We train the adapter by freezing BiDiffuser and minimizing
\begin{equation}
	\begin{aligned}
		\mathcal{L}_\text{ada} = 
  \mathbb{E}_{\bm{\epsilon}^{y},\mathbf{x}_{0}}[
  \Vert
  \bm{\epsilon}^{x} - \bm{\epsilon}_{\theta}^{x}(\mathbf{x}_{t^{x}},y_0,t^{x})\Vert^{2}],
	\end{aligned}
 \label{eq:L_adapter}
\end{equation}
where ${\epsilon}_{\theta}^{x}$ is not updated as BiDiffuser is frozen.

\subsection{Image-to-Text Generation}
\label{sub: i2t}


BiDiffuser can convert images into vectors in the text space, facilitating alignment with the vector space of LLMs. In the following, we show how BiDiffuser can be integrated with LLMs to perform image-to-text generation tasks such as image captioning and visual question answering (VQA).


\subsubsection{Aligning BiDiffuser with LLMs }
\label{visual tuning}

\begin{figure*}[t]
    \centering
    \begin{subfigure}{0.42\linewidth}
        \centering
        \includegraphics[width=\linewidth]{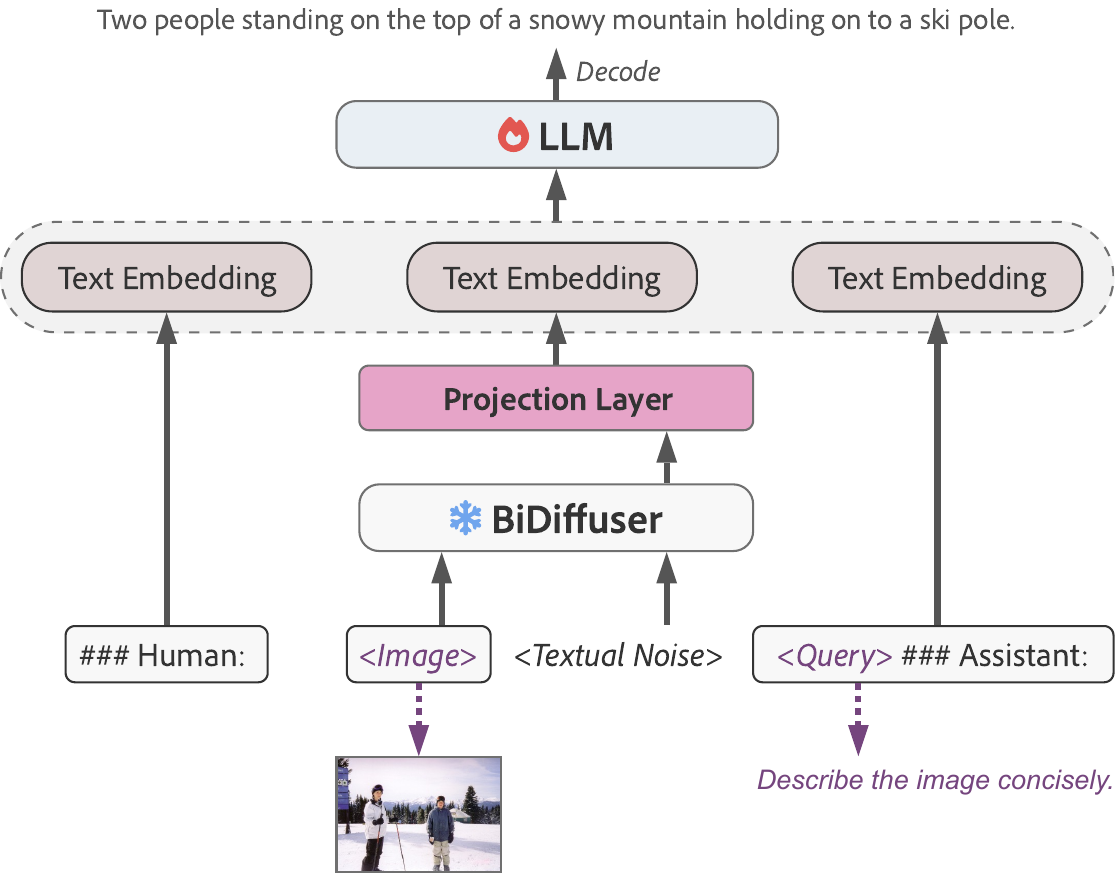}
        \caption{Pre-Align manner.}
        \label{fig:image1}
    \end{subfigure}
    \hspace{2mm} 
    \begin{subfigure}{0.42\linewidth}
        \centering
        \includegraphics[width=\linewidth]{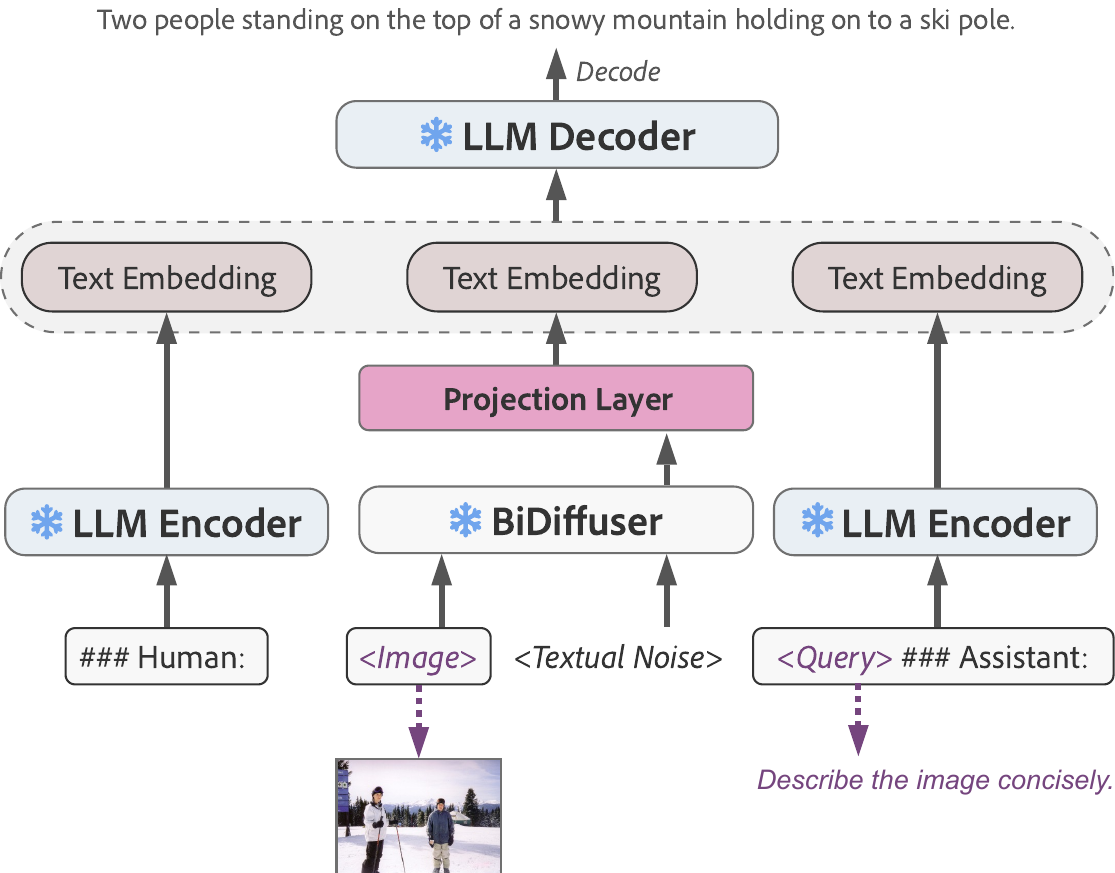}
        \caption{Mid-Align manner.}
        \label{fig:image2}
    \end{subfigure}
    \caption{Two different ways of aligning BiDiffuser with LLMs.
    }
    \label{frame}
\end{figure*}

We connect BiDiffuser and LLMs via a simple projection layer, which maps text embeddings obtained from the output of the diffusion model to the embedding space of LLMs. As shown in Figure~\ref{frame}, the alignment can take place either prior to the LLM (Pre-Align manner) or between its encoder and decoder components (Mid-Align manner). 

%
\textbf{Pre-Align Manner.} As shown in Figure~\ref{fig:image1}, the projection layer is placed before the LLM to map the output of BiDiffuser (image representations) to the text embedding space of the LLM. The text embedding of the input image is then concatenated with the embeddings of the textual instructions and fed to the LLM for decoding. 
To synchronize the text space of BiDiffuser with that of the LLM, we propose to use the image-grounded text generation (ITG) objective to drive the model to generate texts based on the input image by computing the auto-regressive loss:
\begin{equation}
	\begin{aligned}
		\mathcal{L}_\text{ITG} = -\frac{1}{L} \sum_{l=1}^{L}\log p_{\phi}(w_{l}^g|w_{<l}^{g},I,T_I),
	\end{aligned}
        \label{CE_loss}
\end{equation}
where $w^{g}=(w_1^g,...,w_L^g)$ represents the ground-truth caption  of image $I$ with length $L$, $T_I$ is the text instruction, and $\phi$ denotes the model parameters, which include the parameters of the projection layer and the LLM.

\textbf{Mid-Align Manner.} 
As shown in Figure~\ref{fig:image2}, the projection layer is placed between the LLM's encoder and decoder, aiming to map the output of BiDiffuser to the embedding space of the text that is encoded by the LLM's encoder.
Particularly, we argue that the output of BiDiffuser, once mapped by the projection layer and denoted as $\mathbf{d}_{\mathrm{diff}}$, should align with the image caption that is encoded by the LLM's encoder, denoted as $\mathbf{d}_{\mathrm{llm}}$.
Therefore, to accurately learn the alignment between the image and text representations, in addition to the ITG loss in Eq.~\ref{CE_loss}, we also employ an image-text distance minimization (ITDM) loss: 
\begin{equation}
	\begin{aligned}
		\mathcal{L}_\text{ITDM} &= \frac{1}{N}\sum_{i=1}^{N}\|\mathbf{d}_{\mathrm{diff}}-\mathbf{d}_{\mathrm{llm}}\|_{2}^{2}, \\ \mathcal{L}_\text{mid} &= \mathcal{L}_\text{ITG} + \mathcal{L}_\text{ITM}.
	\end{aligned}
\end{equation}
where $N$ is the batch size, and $\mathcal{L}_\text{mid}$ is the overall loss. In this manner, the model parameters $\theta$ only include the parameters of the projection layer.

After the alignment, \modelname{} gains the capability of zero-shot image-to-text generation, including tasks such as image captioning and VQA.


\subsubsection{Instruction-Tuning LLMs 
}
\label{finetuning llm with instructions}


When aligning BiDiffuser with an LLM, we perform instruction-tuning on the LLM to equip it with the capability of understanding multimodal tasks. We designed different instructions for different LLMs, as shown in Table~\ref{tab:response_format_example}. General instruction template is denoted as follows:

\noindent {USER: <Img><image></Img> + Instruction. Assistant: <answer>.}

For the <image> placeholder, we substitute it with the output of BiDiffuser. 
To avoid over fitting to the specific task and counter the model's inclination to generate excessively short outputs, we have devised specific instructions (see Table~\ref{instructionquery}), which enable the LLM to produce concise responses when necessary. 
For different tasks, the distinct instruction templates are as outlined in Appendix~\ref{instruct tuning}.




\subsection{Text-to-Image Response Generation}
\label{text-to-image instruct}

Most of existing multimodal models, including the BLIP series and LLaVA series are unable to provide a multimodal response as they are primarily designed to generate only textual outputs. On the other hand, Emu~\citep{sun2023generative} takes a unified approach to predict the subsequent visual or textual token in an auto-regressive manner, but it is heavily reliant on vast quantities of training data. 
Contrary to the limitations of these existing models, \modelname{}, by leveraging the bidirectional generation capability of BiDiffuser and the inference capability of LLMs, can produce accurate and high-quality visual response with ease.

To tackle multimodal response generation tasks such as PhotoChat~\citep{zang2021photochat}, we first leverage the MLLM to generate detailed image captions based on dialogue context. Then, we employ BiDiffuser to create the corresponding images with the produced captions. Specifically, we replace the image featured in the dialogue with its corresponding descriptive caption, encapsulating it with task-specific tokens <Img>,</Img> and constructing the following instruction templates: 

\noindent {USER: Dialog history. Assistant: <response> + <Img><caption></Img>.}

\begin{figure}[t]
	\begin{center}
	\includegraphics[width=0.45\textwidth]{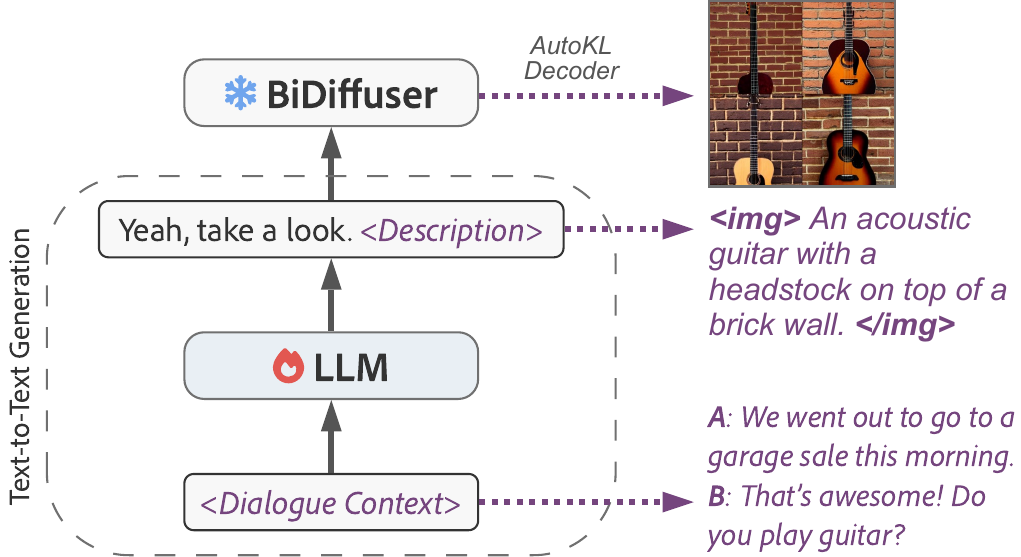}
        \caption{Text-to-image generation by \modelname{}. LLM generates the response and description of the image. BiDiffuser generates images based on the description.}
	\label{frame_revise}
    \end{center}
\end{figure}
\noindent When <caption> appears in response, it represents the generated description of the image. So we can use LLM's original auto-regressive training objective. Specifically, we compute the probability of the target caption by:
\begin{equation}
	\begin{aligned}
            \mathcal{L}_{t2t} = -\frac{1}{L} \sum_{l=1}^{L}\log p_{\vartheta}(w_{l}^c|w_{<l}^{c},H),
	\end{aligned}
 \label{eq:L_t2t}
\end{equation}
where $w^{c}=(w_1^c,...,w_L^c)$ represents the caption of image $x_{0}$ with length $L$, $H$ is the dialogue history, and $\vartheta$ denotes the LLM's parameters.
Considering the potential for alignment deviation in discrete text alone, given the description of the image $\mathbf{y}_{0}$, we utilize $y_{0}$, which is the combination of the SUR adapter's output and the CLIP text encoder's output, as the conditional component of the diffusion model. This directly contributes to the denoising process. The loss function for the denoising process of a noisy image $x_{t^{x}}$ is formulated in a way that is similar to Eq.~\ref{eq:L_adapter}:
\begin{equation}
	\begin{aligned}
		\mathcal{L}_{t2i} &= 
  \mathbb{E}_{\bm{\epsilon}^{y},\mathbf{x}_{0}}[
  \Vert
  \bm{\epsilon}^{x} - \bm{\epsilon}_{\theta}^{x}(\mathbf{x}_{t^{x}},y_0,t^{x})\Vert^{2}],
	\end{aligned}
 \label{eq:L_t2i}
\end{equation}
where ${\epsilon}_{\theta}^{x}$ is not updated and we only train the parameters of LLM and adapter. The overall loss for text-to-image task is:
\begin{equation}
	\begin{aligned}
  \mathcal{L}_{all} &= \mathcal{L}_{t2i} + \mathcal{L}_{t2t}.
	\end{aligned}
 \label{eq:L_final}
\end{equation}
Training with the instruction data enables our model to not only produce text responses but also perform image intent classification and generate image captions that BiDiffuser can interpret. 

\section{Experiments}

\subsection{Experimental Setup}

We initialize encoder-decoder LLM from FlanT5-XL or decoder-only LLM from Vicuna-7B, along with the utilization of the diffusion module from BiDiffuser.
During the alignment process, we maintain the frozen state of the BiDiffuser. 
The statistics of the datasets for pre-training, alignment and instruction-tuning can be found in Appendix~\ref{dataset detail}. For the image captioning task, \modelname{} is evaluated on both the MS-COCO~\citep{lin2014microsoft} Karpathy test set and the NoCaps~\citep{agrawal2019nocaps} validation set.
For the VQA task, we evaluated on OK-VQA~\citep{marino2019ok} validation set and GQA~\citep{hudson2019gqa} test-dev set.


To adapt the model for multimodal dialogue generation, we fine-tune the LLM and projection layers on the PhotoChat dataset. We incorporate photo-sharing activities into the dialogue context by generating <Img><caption></Img>, and utilize cross-entropy loss exclusively for fine-tuning the multimodal generation task.
Given the limited expressiveness of image descriptions in the PhotoChat dataset, as evidenced by Figure~\ref{tab:case_study} in Appendix~\ref{more_examples}, we regenerate image annotations in a text format similar to that used in MS-COCO. 

\begin{table*}
	\centering
	\fontsize{8}{8}\selectfont
	\begin{tabular}{l|cccccccc}
		\toprule
		\multirow{2}{*}{\bf Model}&
		\multicolumn{2}{c}{\bf Dataset Size}&
		\multicolumn{2}{c}{\bf NoCaps (val)}&\multicolumn{2}{c}{\bf COCO (Karpathy)}&\bf OK-VQA&\bf GQA\cr
		\cmidrule(lr){2-3}\cmidrule(lr){4-5}\cmidrule(lr){6-7}\cmidrule(lr){8-8}\cmidrule(lr){9-9}
		&PT&IT&CIDEr&SPICE&BLEU@4&CIDEr&Accuracy&Accuracy\cr
		\midrule
		{ BLIP~\citep{li2022blip} }&129M&-&113.2&14.8&40.4&136.7&-&-\cr
        { Flamingo~\citep{alayrac2022flamingo} }&1.8B&-&-&-&-&138.1&50.6&-\cr
		{ BLIP-2 OPT-6.7B~\citep{li2023blip}}&129M&-&121.0&15.3&\bf 43.5&145.2&36.4&36.4\cr
		{ BLIP-2 FlanT5XL~\citep{li2023blip}}&129M&-&121.6&\bf 15.8&42.4&144.5&39.4&44.4\cr
            { InstructBlip 7B~\citep{instructblip}}&129M&1.2M&\bf 123.1&-&40.8&140.7&$61.0^{\star}$&$49.2^{\star}$\cr
             { MiniGPT-4~\citep{zhu2023minigpt}}&-&5M&42.4&-&-&-&37.5&30.8\cr
             { LLaVA~\citep{liu2023visual}}&558K&158K&33.1&-&7.9&30.0&54.4&41.3\cr
		\cmidrule(lr){1-9}
		{\bf \modelname{} FlanT5XL}&169K&90K&{121.2}&{15.5}&{\bf 43.5}&{\bf 145.7}&41.1&37.2\cr
        {\bf \modelname{} Vicuna-7B}&169K&90K&{121.8}&{\bf 15.8}&{42.4}&{144.6}&45.2&\bf 44.6\cr
		\bottomrule
	\end{tabular}
 	\caption{Evaluations of \modelname{} and baselines on various \textbf{image understanding} tasks. PT, IT indicate sample sizes in the pretraining and instruction tuning stages respectively. \modelname{}'s results on NoCaps, OK-VQA and GQA were obtained in a zero-shot setting. $\star$ denotes that the model was trained on other VQA datasets. }
        \label{tab:performance_comparison}
\end{table*}

\begin{table}[t]
	\centering
	\fontsize{7.2}{8}\selectfont
	\begin{tabular}{l|cccccccccc}
		\toprule
		\multirow{2}{*}{\bf Model}
  &
		\multicolumn{3}{c}{\bf Response Generation}&{\bf Image }\cr
		\cmidrule(lr){2-4}\cmidrule(lr){5-5}
		&BLEU-1/2&PPL$\downarrow$&ROUGE-L&FID$\downarrow$\cr
		\midrule
		{Divter~\citeauthor{sun2021multimodal}}&6.5/1.7&59.6&5.69&29.16\cr
		Maria~\citeauthor{liang2021maria}&13.8/9.2&48.7&15.17&-\cr
        \specialrule{0.05em}{0.3em}{0.3em}
		{\bf \modelname{} FlanT5}&{22.3/18.7}&{13.3}&{17.24}&{ 10.30}\cr
            {\bf \modelname{} Vicuan}&{\bf 23.6/19.9}&{\bf 11.3}&{\bf 18.85}&{\bf 9.72}\cr
            + w/o adapter &{\bf -}&{-}&{-}&{ 10.16}\cr
	\bottomrule
	\end{tabular}
        \caption{Evaluation on the \textbf{PhotoChat} dataset.}
        \label{tab:performance_comparison_photo}
\end{table}

\begin{table}
  \begin{minipage}{0.99\linewidth}
\centering
\fontsize{8.5}{8}\selectfont
\scalebox{0.95}{
\begin{tabular}{l|c|c|c }
\toprule
 MLLM & Sample Size	& Cosine Similarity $\uparrow$ & MSE $\downarrow$ \cr
\midrule
MiniGPT-4 &5M&0.0016&6.2031 \cr
LLaVA v1.5 &558K&-0.0026&0.8433 \cr
Emu &2B&0.0054&0.4062 \cr
\modelname{} &169K&0.0128&0.0338\cr
\bottomrule
\end{tabular}
}
\captionof{table}{\textbf{Data efficiency.} Avg. Cosine similarity and mean square error between the projected representations and their respective captions embedded by LLM.}
\label{tab:representation distance}  
  \end{minipage}
\end{table}

\begin{table*}[ht]
	\centering
	\fontsize{8.5}{8}\selectfont
 
	\begin{tabular}{ccc|cccccc}
\toprule
\multirow{2}{*}{\textbf{LLM}}&\multirow{2}{*}{\textbf{\shortstack {Diffusion\\ Model}}}& \multirow{2}{*}{\textbf{Alignment}} &
\bf NoCaps &\multicolumn{3}{c}{\bf COCO(Karpathy)}&\bf OK-VQA\cr
\cmidrule(lr){4-4}\cmidrule(lr){5-7}\cmidrule(lr){8-8}
&& &CIDEr&SPICE&BLEU@4&CIDEr&Accuracy\cr
\midrule
\freeze{} T5 & UniDiffuser & Pre-Align
&62.4&{18.0}&{26.8}&90.7&33.0\cr
\tune{} T5 & BiDiffuser & Pre-Align
&119.1&{\bf 25.5}&{42.6}&145.1&{\bf 41.1}\cr
\freeze{} T5 & BiDiffuser & Mid-Align
&121.2&25.1&43.5&{\bf 145.7}&31.5\cr
\tune{} T5 & BiDiffuser & Mid-Align
&121.5&25.3&\bf 43.6&{\bf 145.7}&36.4\cr
\midrule
\tune{} Vicuna-7B & BiDiffuser & Pre-Align
&{\bf 121.8}&24.9&42.4&144.6&{\bf 45.2}\cr
\freeze{} Vicuna-7B & BiDiffuser & Pre-Align
&119.0&24.6&40.3&140.3&{42.7}\cr
\bottomrule
	\end{tabular}
 	\caption{\textbf{Ablation study} on image captioning and VQA tasks. \tune{} / \freeze{} denotes tuning/freezing the LLM.
  }
    \label{tab:ablation study}
\end{table*}

\subsection{Evaluation}

We evaluate \modelname{} on various vision-language tasks including image captioning (MS-COCO~\citep{lin2014microsoft}, NoCaps~\citep{agrawal2019nocaps}), visual question answering (OK-VQA~\citep{marino2019ok}, GQA~\citep{hudson2019gqa}), and multimodal dialog generation (PhotoChat~\citep{zang2021photochat}). We use BLIP~\citep{li2022blip}, Flamingo~\citep{alayrac2022flamingo}, BLIP-2~\citep{li2023blip}, InstructBlip~\citep{instructblip}, MiniGPT-4~\citep{zhu2023minigpt}, and LLaVA~\citep{liu2023visual} as baselines for image-to-text tasks, and Maria~\citep{liang2021maria} and Divter~\citep{sun2021multimodal} as baselines for the multimodal response generation task. See details in Appendix~\ref{baselines} and ~\ref{implementation details}.

\subsection{Overall Results}

Tab.~\ref{tab:performance_comparison} presents the evaluation results for each baseline and our models on MS-COCO and VQA (zero-shot) datasets. \modelname{} outperforms most of the baseline models on both the COCO test set and NoCaps validation set (zero-shot transfer). Despite being pre-trained on a small dataset (MS-COCO), \modelname{}'s performance on the image captioning generation task is comparable to models trained on larger datasets. Additionally, on the OK-VQA and GQA datasets, \modelname{} demonstrates improved performance compared to other models of a similar scale, achieving higher accuracy even with a simple greedy search decoding method.

In Tab.~\ref{tab:performance_comparison_photo}, the evaluation results on the PhotoChat dataset are presented. Our method exhibits clear advantages in terms of PPL, indicating strong performance on response generation task. Because of the image descriptions in the PhotoChat dataset are overly concise, we utilized \modelname{} to regenerate the image descriptions, which improved the performance of our model on image generation compared to other models. Additionally, with the adapter, \modelname{} is capable of generating images of superior quality.

\subsection{Ablation Study}
\label{ablation detail}

In Tab.~\ref{tab:ablation study}, we examine the impact of freezing/tuning BiDiffuser and the LLM.
It can be observed that frozen Mid-Align method outperforms Pre-Align method in image captioning, which shows ITDM loss function is effective. However, the frozen Mid-Align method exhibits inferior performance in the VQA task. We hypothesize that this is due to the integration of mid-aligned target image features with query information, and the projection layer is insensitive to instruction information. We conduct instruction-tuning on Pre-Align T5 and Vicuna. Compared to models at the same scale, these instruction-tuned models achieve superior results.

\begin{table*}
	\centering
	\fontsize{8.5}{8}\selectfont
	\begin{tabular}{l|c|c|c|c}
		\toprule
  
          Model&IT&VQAv2 (test-dev)&TextVQA&MMBench (test)\cr
          \midrule
         MiniGPT-4~\citep{zhu2023minigpt}&5M&-&19.4&23.0\cr
         InstructBLIP Vicuna-7B~\citep{instructblip}&1.2M&-&50.1&33.9\cr
         LLaVA-1.5 Vicuna-7B~\citep{liu2023improved}&665K&78.5&58.2&65.2\cr
         LLaVA-1.5 Vicuna-13B~\citep{liu2023improved}&665K&80.0&61.3&67.8\cr
         \rowcolor{mygray} \modelname{} Vicuna-7B w/ ViT-L&251K&79.4&57.9&63.9\cr
         \rowcolor{mygray} LLaVA-1.5 Vicuna-7B w/ \modelname{}&665K&80.2&58.8&66.1\cr
         \rowcolor{mygray} LLaVA-1.5 Vicuna-13B w/ \modelname{}&665K&\bf 80.5&\bf 61.5&\bf 69.2\cr
		\bottomrule
	\end{tabular}
        \caption{Evaluation of \textbf{\modelname{} variants} and baselines on more complex VQA tasks and the latest MMBench benchmark. ``w/ \modelname{}'' means incorporating the core components of our model into existing models as depicted in Figure~\ref{vqa_tune} in Appendix~\ref{instruct tuning}. \modelname{} variants rank among the top models on the leaderboard of MMBench.
        }
        \label{VQA results}
\end{table*}


\subsection{Data Efficiency in Training}
\label{data_efficiency}



In Tab.~\ref{tab:representation distance}, we examine the data efficiency of different image encoders for alignment with LLMs. \modelname{} uses BiDiffuser, which maps images to the text space, simplifying alignment with LLMs. To assess the quality of visual representations, we measured the distance between the projected representations and their respective captions embedded by an LLM. We randomly selected 1,000 images with their corresponding captions from the MSCOCO dataset. The results show that our model, EasyGen, aligns significantly better with the LLM compared to other CLIP-based MLLMs, despite using less data for alignment. This indicates the effectiveness of our approach in achieving strong alignment with LLMs.

\begin{table}
  \begin{minipage}{0.99\linewidth}
\centering
\fontsize{8.5}{8}\selectfont
\scalebox{0.95}{
\begin{tabular}{l|ccc }
\toprule
 MM-Model & FID $\downarrow$ & Diffusion Model & FID $\downarrow$ \cr
\midrule
\multicolumn{3}{l}{\emph{Zero-Shot}} \cr
NExT-GPT &11.28 (+0.07)&SD&11.21 \cr
Emu &11.66 (+1.73)&SD v1.5&9.93 \cr
\modelname{} &9.16 (-0.55)&UniDiffuser&9.71\cr
+ w/o adapter &9.52 (-0.19)&UniDiffuser&9.71\cr
\midrule
\multicolumn{3}{l}{\emph{Fine-tuned on MS-COCO}} \cr
\modelname{} &7.68 (-0.44)&UniDiffuser&8.12\cr
+ w/o adapter &7.89 (-0.23)&UniDiffuser&8.12\cr
\bottomrule
\end{tabular}
}
\captionof{table}{Comparing the \textbf{image generation quality} of end-to-end MLLMs and their corresponding diffusion models on the MS-COCO validation set (256 × 256). Our \modelname{} surpasses the original diffusion model, while other MLLMs fall short in comparison.  }
\label{tab:fid distance}  
  \end{minipage}
\end{table}

\subsection{Image Generation Quality}
\label{sec:t2i}


Tab.~\ref{tab:fid distance} evaluates the generated image's quality of MLLMs on MS-COCO validation set, using 30K randomly selected prompts to compute the FID score on generated images. To confirm the efficacy of our approach, we fine-tuned our method on a portion of the original data (LIAON-COCO) and the MS-COCO train set, respectively. While other models resulted in a decrease in image generation performance compared to the corresponding diffusion model, \modelname{} outperformed the original UniDiffuser due to the fine-tuned BiDiffuser and the adapter module.  Furthermore, Tab.~\ref{tab:clip-t score} provides CLIP-T scores from ImagenHub. We notice similar trends to the results in Tab.~\ref{tab:fid distance} using the FID indicator. This suggests that our method can better align LLM with diffusion model's text space.

\begin{table}
  \begin{minipage}{0.99\linewidth}
\centering
\fontsize{8.5}{8}\selectfont
\scalebox{0.95}{
\begin{tabular}{l|ccc }
\toprule
 MM-Model & CLIP-T~$\uparrow$ & Diffusion Model & CLIP-T~$\uparrow$ \cr
\midrule
NExT-GPT &0.259 (-0.031)&SD&0.290 \cr
Emu &0.262 (-0.023)&SD v1.5&0.285 \cr
Emu2 &0.266 (-0.023)&SD XL&0.289 \cr
\modelname{} &9.16 (-0.55)&UniDiffuser&9.71\cr
\bottomrule
\end{tabular}
}
\captionof{table}{Comparing the \textbf{CLIP-T score} of end-to-end MLLMs and their corresponding diffusion models on the ImagenHub.}
\label{tab:clip-t score}  
  \end{minipage}
\end{table}

\subsection{Extendability}
\label{integrating MLLM}
Tab.~\ref{VQA results} explores the extensibility of our method from two aspects. Firstly, we aim to enhance the performance of \modelname{} on complex tasks such as VQA and OCR by integrating more powerful visual encoders. Considering the potential information dilution or omission when using BiDiffuser to convert images to text space, we choose to integrate CLIP ViT-L/14 as the image encoder (as depicted in Figure~\ref{vqa_tune} in the Appendix). During this process, we freeze CLIP and BiDiffuser while fine-tuning the parameters of the LLM and projection layers. The results presented in Tab.~\ref{VQA results} include performance on traditional short QA and the modern benchmark MMBench~\cite{liu2023mmbench}. With CLIP ViT-L, \modelname{}'s performance is better than LLaVA on the VQAv2 dataset, demonstrating that BiDiffuser can effectively assist LLM in understanding images.
Secondly, we investigate the plug-and-play capability of BiDiffuser, as it can also be integrated into other MLLMs (with the same LLMs) to improve their performance. The results demonstrate that with BiDiffuser, LLaVA-1.5 could achieve better performance. We speculate that BiDiffuser provides guidance information to MLLMs, enabling them to better understand the details of CLIP encoded images.

\section{Conclusion}

We have introduced \modelname{}, a model that facilitates multimodal understanding and generation. Compared to existing models, \modelname{} offers a more efficient solution by employing BiDiffuser, a bidirectional diffusion model. This allows for more effective modal interactions, handling both image-to-text and text-to-image generations by the fusion of BiDiffuser and LLMs. Additionally, EasyGen can be easily integrated into existing advanced multimodal LLMs to further boost their performance.



\section{Limitations}
This section aims to highlight the limitations of our work and provide further insights into the research in this area. Our model relies on diffusion for multi-modal interaction, which means that the text-to-image and image-to-text processes may take longer. In our experiments, we tested the performance of our model on one A100 (80G) GPU. During inference, using 1000 image-caption pairs, \modelname{} took approximately 2.95 seconds for the caption generation task (with the diffusion module taking about 2.41 seconds) and around 4.96 seconds to generate an image. We believe it would be beneficial to explore more efficient sampling methods, such as DPM-Solver++~\cite{lu2022dpm}, to improve the overall efficiency of \modelname{}.


\bibliography{custom}

\appendix

\clearpage

\section{Ethics Statement}
We adhere to the ACL Ethics Policy and have conducted our research using publicly available repositories and datasets. Our primary focus is on investigating the integration of diffusion models and LLMs for multimodal generation. Therefore, the results should be seen as AI-generated content. While we have not observed deliberate harmful content, the model has the potential to generate such content if triggered. We have taken steps to minimize this risk through fine-tuning on public datasets, but caution is still exercised. In future, we will prioritize improving downstream performance and exploring methods to enhance control over the generation process. To ensure reproducibility and support future research, we have made all resources publicly available and provided proper citations to previous research within the code.
\section{Datasets}
\label{dataset detail}
We test the effectiveness of \modelname{} by experimenting on different tasks including image captioning, visual question answering (VQA), and multimodal dialogue tasks. Table~\ref{VQA datasets} shows the statistics of the pre-training datasets for BiDiffuser, alignment and VQA tasks.


We use the MS-COCO~\citep{lin2014microsoft} dataset for image captioning. Following BLIP-2~\citep{li2023blip}, we fine-tune \modelname{} on MS-COCO and evaluate its performance on the Karpathy test set and the NoCaps~\citep{agrawal2019nocaps} validation set. In MS-COCO, each image typically has five captions that convey similar meanings. The training set consists of 82,783 images with 414,113 captions, while the COCO Karpathy test set has 5,000 images and the NoCaps validation set has 4,500 images.



For multimodal dialogue, we utilize the PhotoChat~\citep{zang2021photochat} dataset, which is a high-quality dataset consisting of 10,917 images and 12,286 dialogues. Each dialogue is associated with a user image and its corresponding text description. The dataset is divided into 10,286 training instances, 1,000 development instances, and 1,000 testing instances.
Moreover, PhotoChat includes photo-sharing activities, defined as the process of creating <Img><caption></Img> in this study. Each conversation in PhotoChat is broken down and constructed into multiple samples so that each round of responses can be learned. Specifically, we regard the first three turns as the dialog context, and the subsequent turns as the prediction targets. By converting the dialogues of this dataset into the form mentioned in \ref{text-to-image instruct}, we obtained 49,240 train, 4,792 dev, and 4,836 test dialogue pairs.

For the VQA task, we conduct a quantitative evaluation on both the OK-VQA~\citep{marino2019ok} validation set (5,046 questions) and the GQA~\citep{hudson2019gqa} test-dev set (12,578 questions).
As shown in Table~\ref{tab:ablation study}, for the frozen LLM, following BLIP-2, we employ the length penalty in beam search to encourage short answer generation. On the contrary, for the tuned LLM, we use the VQA instructions (as shown in Table~\ref{instruction templates}) to do instruction tuning during the alignment process. 
The data for instruction tuning is constructed by randomly selecting 5K data from VQAv2~\citep{goyal2017making} and 5K data from Visual Dialog~\citep{visdial_diversity} training set.

\begin{table*}[ht]
	\centering
	\fontsize{7}{8}\selectfont

	\begin{tabular}{l|l|c|c|c|c}
		\toprule
         Data types&Dataset&Size&BiDiffuser&Alignment&Fine-tuning\cr
          \midrule
         \multirow{2}{*}{Caption}&MS-COCO caption~\citep{lin2014microsoft} & 83K & \CheckmarkBold & \CheckmarkBold & \XSolidBrush \cr
          &Visual Genome~\citep{krishna2017visual} & 86K & \CheckmarkBold &  \XSolidBrush & \XSolidBrush \cr
         \midrule
         Multimodal instruction & LLaVA dataset~\cite{liu2023visual} & 80K & \XSolidBrush & \CheckmarkBold & \CheckmarkBold \cr
         \midrule
         \multirow{2}{*}{VQA}&  VQAv2~\citep{goyal2017making} & 83K & \XSolidBrush & - & \CheckmarkBold \cr
         &AOK-VQA~\citep{schwenk2022okvqa} & 66K & \XSolidBrush & \XSolidBrush & \CheckmarkBold \cr
         \midrule
         {\multirow{2}{*}{OCR-related tasks}}&Text Captions~\citep{sidorov2020textcaps} & {\multirow{2}{*}{22K}} & \XSolidBrush & \XSolidBrush & \CheckmarkBold \cr
         & TextVQA~\citep{Singh_2019_CVPR} & & \XSolidBrush & \XSolidBrush & \CheckmarkBold \cr
		\bottomrule
	\end{tabular}
        \caption{ Description of datasets used in our alignment and VQA fine-tuning stages. Noting that in alignment process, we used 5K images from VQAv2 dataset.}
        \label{VQA datasets}
\end{table*}

\begin{table*}[htbp]
	\centering
	\fontsize{7.8}{8}\selectfont
	\begin{tabular}{c|lccc}
		\toprule
          &\textbf{Dataset}&\textbf{Task}&\textbf{Split}&\textbf{Metric}\cr
          \midrule
         \multirow{4}{*}{{\textbf{Image-to-Text}}} & MS-COCO~\citep{lin2014microsoft} &Image captioning&Test& CIDEr, BLEU, SPICE\cr
         \multirow{4}{*}{} & NoCaps~\citep{agrawal2019nocaps} &Image captioning&Val&CIDEr, SPICE\cr
         \multirow{4}{*}{} & OK-VQA~\citep{marino2019ok} &VQA&Val&Accuracy\cr
         \multirow{4}{*}{} & GQA~\citep{hudson2019gqa} &VQA&Test&Accuracy\cr
         \midrule
         \textbf{Multimodal Generation}& PhotoChat~\citealp{zang2021photochat} &Image dialogue&Test&PPL, BLEU, ROUGE, FID\cr
		\bottomrule
	\end{tabular}
        \caption{Summary of the evaluation datasets and metrics.}
        \label{datasets summary}
\end{table*}

\begin{table*}[htbp]
	\centering
	\fontsize{8}{8}\selectfont
	\begin{tabular}{l|l}
		\toprule
          \textbf{Task}&\textbf{Instruction Template}\cr
          \midrule
         \textbf{Image Captioning}&USER: <image>+random[query] Assistant:\cr
         \midrule
         \multirow{3}{*}{\textbf{LLaVA 80K}} & USER: Please answer question from this image: <image> Question: <question> Assistant:\cr
         \multirow{3}{*}{} & USER: Image: <image> Question: <question> Assistant:\cr
         \multirow{3}{*}{} & USER: Answer question <question> through the image <image> Assistant:\cr
         \midrule
         \textbf{Multimodal Dialogue}&USER: Dialog history+<photo>+Dialogue history Assistant:\cr
         \midrule
         \multirow{2}{*}{\textbf{VQA}} &USER: Image: <image> Question: <question> {Short answer:} Assistant:\cr
        \multirow{2}{*}{} & USER: Image: <image> Question: <question> Answer the option’s letter. Assistant:\cr
		\bottomrule
	\end{tabular}
        \caption{Examples of task instruction templates. 
        <image> represents the input image, <question> denotes the question in the VQA and LLaVA 80K dataset, and <photo> is the image description of the input image.}
        \label{instruction templates}
\end{table*}

\section{Baselines}
\label{baselines}

We compare our proposed model with the following state-of-the-art baselines:

\noindent\textbf{BLIP}~\citep{li2022blip} is a multimodal mixture of encoder-decoder. It can be used as an image-based text encoder or decoder. We use it to perform caption generation and VQA tasks.

\noindent\textbf{BLIP-2}~\citep{li2023blip} is pre-trained through bootstrapped learning from frozen visual encoder and LLMs using an efficient pre-training strategy. 

\noindent\textbf{Flamingo}~\citep{alayrac2022flamingo} incorporates new cross-attention layers into Chinchilla language model~\citep{hoffmann2022training} to inject visual features, and pre-trains the new layers on billions of image-text pairs. We use it to perform caption generation and VQA tasks.

\noindent\textbf{InstructBlip}~\citep{instructblip} is a vision-language instruction tuning framework that is trained with BLIP-2 and capable of solving various visual language tasks.

\noindent\textbf{MiniGPT-4}~\citep{zhu2023minigpt} utilizes a single projection layer to align visual information from a pre-trained vision encoder with an LLM. It employed the same visual encoder as used in BLIP-2.

\noindent\textbf{LLaVA}~\citep{liu2023visual} employs a solitary projection layer to convert image features extracted from the pre-trained CLIP-ViT-L/14 visual encoder into the language embedding space of Vicuna.

\noindent\textbf{Maria}~\citep{liang2021maria} is a neural conversation agent which can leverage visual world experiences sourced from a vast image index. It possesses the ability to fetch a relevant image specific to the conversation and extract visual knowledge from it.

\noindent\textbf{Divter}~\citep{sun2021multimodal} focuses on exploring multimodal dialogue generative models. Given the dialogue context, this model first generates a text response or image description and then generates an image according to the description.


\begin{figure}[ht]
	\begin{center}
	\includegraphics[width=0.48\textwidth]{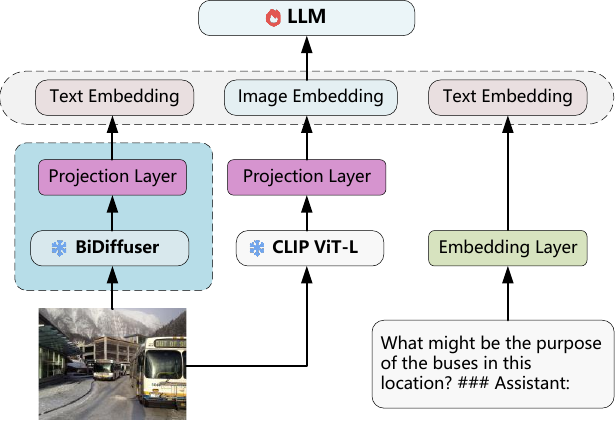}
        \caption{Model's architecture for VQA finetuning. The module with blue background is referred to as BiDiffuser, while the rest is the architecture of MLLM using CLIP as the image encoder (such as LLaVA).
        }
	\label{vqa_tune}
    \end{center}
\end{figure}

\section{Evaluation}
\label{evaluation}

For evaluating the quality of text generation, we utilize metrics such as BLEU, Rouge-L, Accuracy, and PPL (Perplexity). Additionally, following the approach of Vicuna~\citep{vicuna2023} and LLaVA~\citep{liu2023visual}, we employ ChatGPT to assess the generated responses from our model. Specifically, for the image captioning task, we randomly select 30 images from the MS-COCO Karpathy split and then let ChatGPT score the responses generated by \modelname{} and the baseline models. ChatGPT evaluates the models' responses based on relevance, details, and accuracy and assigns an overall score between 1 and 10, with a higher score indicating better performance.
To evaluate the quality of image generation, we use the Frechet Inception Distance (FID) score~\citep{heusel2017gans}, which measures the divergence between two multivariate normal distributions.

\section{Implementation Details}
\label{implementation details}


\paragraph{LLM} During the alignment process, we utilize the AdamW optimizer with $\beta_{0}$ = 0.9, $\beta_{1}$ = 0.99, and weight decay of 0. The LLMs are trained with a cosine learning rate of 2e-5 and a warmup rate of 0.03. We use a batch size of 96 for the frozen LLMs and 32 for the tuned LLMs. During training, we convert the LLMs (FlanT5XL/Vicuna-7B) to BFloat16/FP16 and BiDiffuser to FP16. During the VQA tuning process, we use CLIP ViT-L/14 336px as additional image encoder. We finetune \modelname{} on mixture datasets for 1 epoch with a batch size of 32. We adopt the AdamW optimizer with $\beta$ = (0.9, 0.99) with the learning rate is 2e-5. We use a cosine learning rate decay with a learning rate is 2e-5 and warmup ration is 0.03.

\paragraph{Diffusion Module} We inherit the settings from UniDiffuser and utilize pre-trained weights from its checkpoint for our text-to-image generator. The model is fine-tuned on the MS-COCO and VG dataset, which contains images with a resolution of $512 \times 512$, for 10 epochs with a batch size of 312. For all of our sampling processes, we employ DPM-Solver with 50 steps.



\begin{table}[htbp]
	\centering
	\fontsize{8}{8}\selectfont
	\begin{tabular}{l}
		\toprule
         1. Describe the image \textcolor{blue}{concisely}.\cr
         2. Provide a \textcolor{blue}{brief} description of the given image.\cr
         3. Can you describe this image \textcolor{blue}{briefly}?\cr
         4. Provide a \textcolor{blue}{summary} of visual elements depicted in the image.\cr
         5. Give me the essential characteristics of the photograph in a \\ \textcolor{blue}{concise} manner.\cr
         6. Rephrase the image depicted in a \textcolor{blue}{concise} manner.\cr
         7. Describe the objects in this image \textcolor{blue}{no in detail}.\cr
         8. Please introduce the image for me \textcolor{blue}{briefly}.\cr
         9. Give me the image's short descriptions.\cr
         10. Please provide a \textcolor{blue}{general} depiction of the image presented.\cr
		\bottomrule
	\end{tabular}
        \caption{For the image captioning task, a query instruction is randomly selected.}
        \label{instructionquery}
\end{table}

  

\section{Instruction Tuning}
\label{instruct tuning}
We list the instructions for different tasks in the main paper in Table~\ref{instruction templates}. Specifically, the queries used to describe image contents are presented in Table~\ref{instructionquery}.  Table~\ref{instruction templates} shows the templates used in Vicuna, if the LLM is FlanT5, kindly use ``Human'' to substitute ``USER'' in the instruction templates.
Model architecture for VQA finetuning is shown in Figure~\ref{vqa_tune}. \modelname{} integrates the outputs of BiDiffuser with images encoded by CLIP ViT-L/14. We freeze CLIP and BiDiffuser while only tuning the parameters of the LLM and projection layers.

\begin{table}
  \begin{minipage}{0.99\linewidth}
\centering
\fontsize{8.5}{8}\selectfont
\scalebox{0.95}{
\begin{tabular}{p{1.5cm} p{5.5cm} }
\toprule
 \multicolumn{2}{l}{\bf Different Instruction Templates:}  \\
\midrule
&  \includegraphics[height=2.5cm]{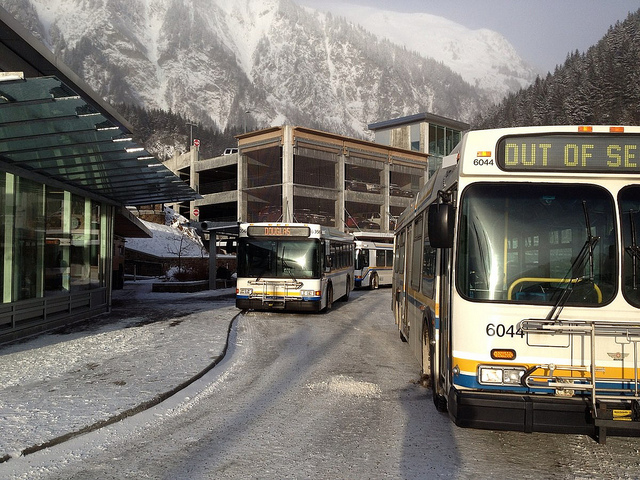} \\

Caption Generation & {\bf Provide a brief description of the given image. Assistant:} \\
Response & Buses parked with a snow mountain view behind them.
\\
\midrule
VQA & What numbers are displayed on the front of the bus on the right? {\bf Short answer: Assistant: } \\
Response & 6044 \\
\midrule
Multimodal Dialogue & What might be the purpose of the buses in this location? {\bf Assistant:}\\
Response & It is plausible that they are in this location for multiple reasons. Some possible reasons might be: 1)...2)... \\
\bottomrule
\end{tabular}
}
\captionof{table}{Examples of different instructions with different output formats. We use bold fonts to indicate different instructions.}
\label{tab:response_format_example}  
  \end{minipage}
\end{table}

\section{Training Efficiency}
\label{efficiency}
Table~\ref{training cost} summarizes the key factors in training \modelname{}. The training process of \modelname{} is computationally efficient, especially with the utilization of the parameter-efficient fine-tuning method LoRa~\citep{hu2021lora}.
To enable multimodal response generation, we further train the aligned \modelname{}. This process entails fine-tuning the LLM (FlanT5XL) on the PhotoChat dataset for 2 epochs, which typically requires approximately 4 A100 (80G) GPU hours.

\section{Impact of Alignment Manners}
In Table~\ref{tab:performance_strategy}, we investigate the impact of different alignment manners on \modelname{}. After removing the ITDM loss, the performance is slightly weaker than the original model. It is evident that the MSE Loss can help to align the semantic spaces of the two models. Furthermore, the performance of the model will drop significantly after removing the cross-entropy loss, suggesting that constraints via the language model play a key role.

\begin{table*}[htbp]
	\centering
	\fontsize{8}{7}\selectfont
	\begin{tabular}{lccc}
		\toprule
          \textbf{Model}&\textbf{Trainable Param.}&\textbf{Training Images}&\textbf{Training Cost}\cr
          \midrule
         {\emph{Pre-training}} \cr
         {\quad BiDiffuser}&952M&169K&120 (A100 80GB) GPU hours\cr
         \midrule
         {\emph{Alignment}} \cr
         {\quad Projection Layers + \freeze{} T5XL}&4M&163K&20 (RTX3090 24GB) GPU hours\cr
         {\quad Projection Layers + \tune{} T5XL}&3B&173K&20 (A100 80GB) GPU hours\cr
         {\quad Projection Layers + \tune{} Vicuna 7B}&7B&173K&72 (A100 80GB) GPU hours\cr
         {\quad Projection Layers + \tune{} Vicuna 7B(LoRa)}&610M&173K&20 (A100 80GB) GPU hours\cr
	\bottomrule
	\end{tabular}
        \caption{ \modelname{}'s trainable parameters, training data size, and training cost during alignment process.}
        \label{training cost}
\end{table*}

\begin{table*}[htbp]
	\centering
	\fontsize{8}{8}\selectfont
	\begin{tabular}{l|cccccccc}
		\toprule
        \multirow{2}{*}{\bf Model}
  &
		\multicolumn{2}{c}{\bf NoCaps (val)}&\multicolumn{3}{c}{\bf COCO (Karpathy)}&\bf OK-VQA&\bf GQA \cr
		\cmidrule(lr){2-3}\cmidrule(lr){4-6}\cmidrule(lr){7-7}\cmidrule(lr){8-8}
		&CIDEr&SPICE&SPICE&BLEU@4&CIDEr&Accuracy&Accuracy\cr
		\midrule
		{\bf \modelname{} Mid-Align FlanT5XL}&{121.2}&{\bf 15.5}&\bf 25.1&{\bf 43.5}&{\bf 145.7}&31.5&22.6\cr
		{+ w/o ITDM}&{118.6}&{15.3}&24.8&{42.2}&141.5&-&-\cr
		{+ w/o ITG}&{93.2}&{12.9}&23.0&{35.1}&127.6&-&-\cr
		\bottomrule
	\end{tabular}
 	\caption{Ablation studies on the instruction-tuning process
and loss functions.}
        \label{tab:performance_strategy}
\end{table*}

\begin{figure*}[t]
    \centering
    \includegraphics[width=0.95\linewidth]{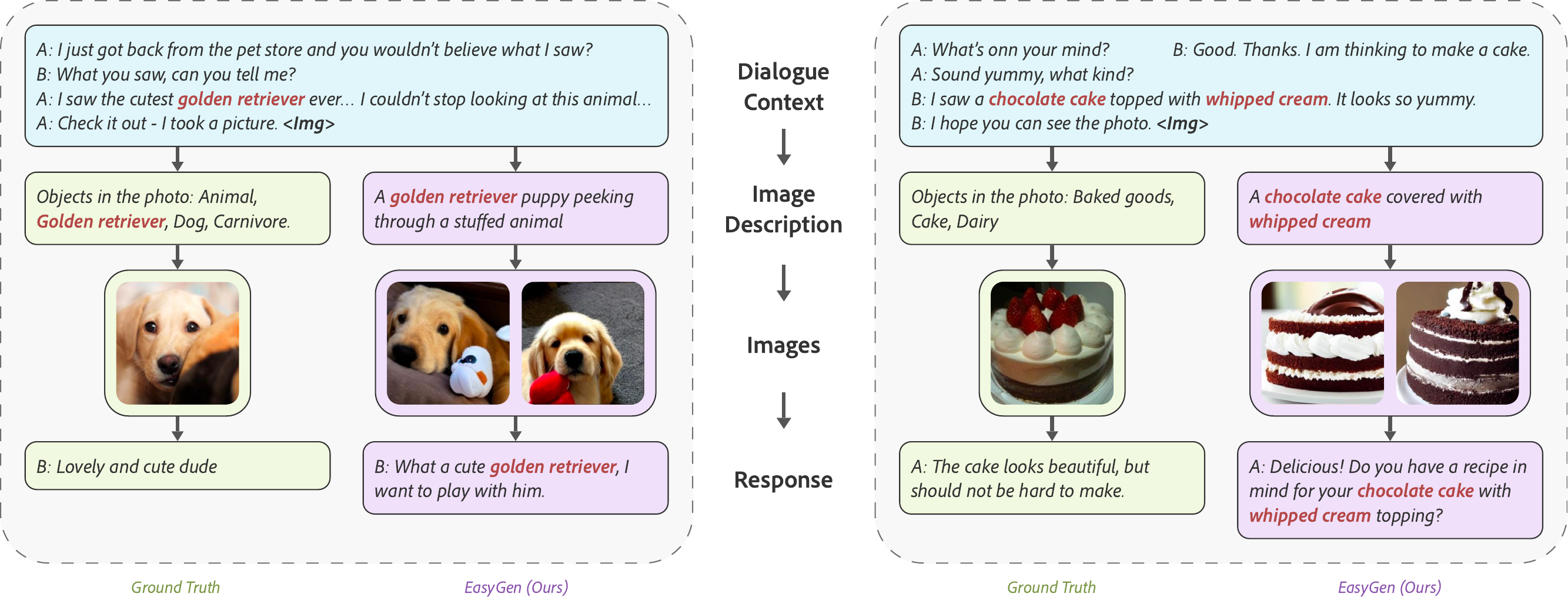}
    \caption{Examples of the generated responses on PhotoChat dataset. The text highlighted in red indicates the objects present in the image. The turns prefixed with A/B denote the given context.
    }
    \label{tab:case_study}
\end{figure*}

\section{More Qualitative Results}
\label{more_examples}
\begin{figure*}[htbp]
	\centering 
	\includegraphics[width=12.5cm, scale=1]{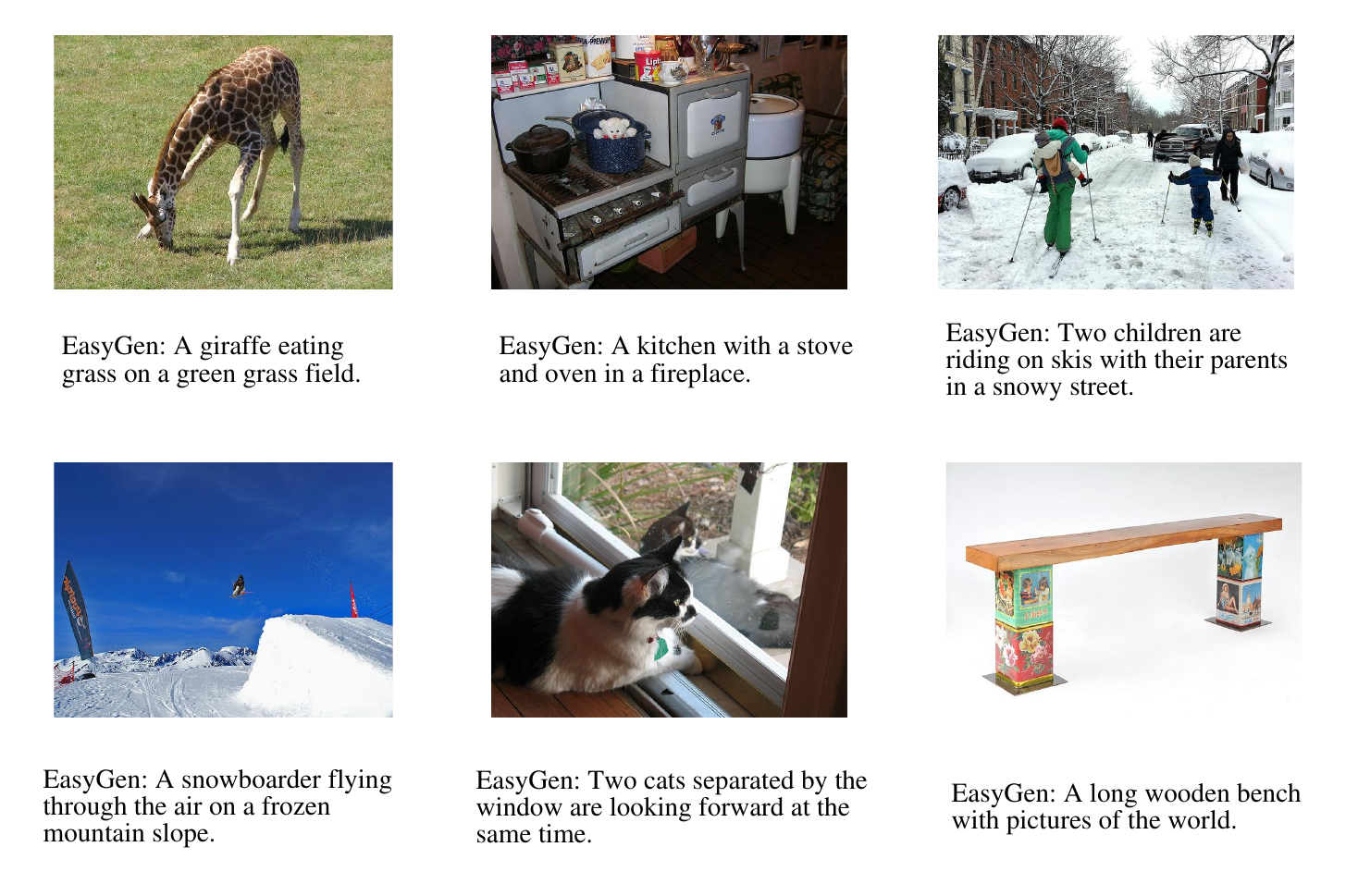} 
	\caption{Examples of image captioning results by \modelname{}.}
        \label{example1}
\end{figure*}

\begin{figure*}[htbp]
	\centering
	\includegraphics[width=12.5cm, scale=0.8]{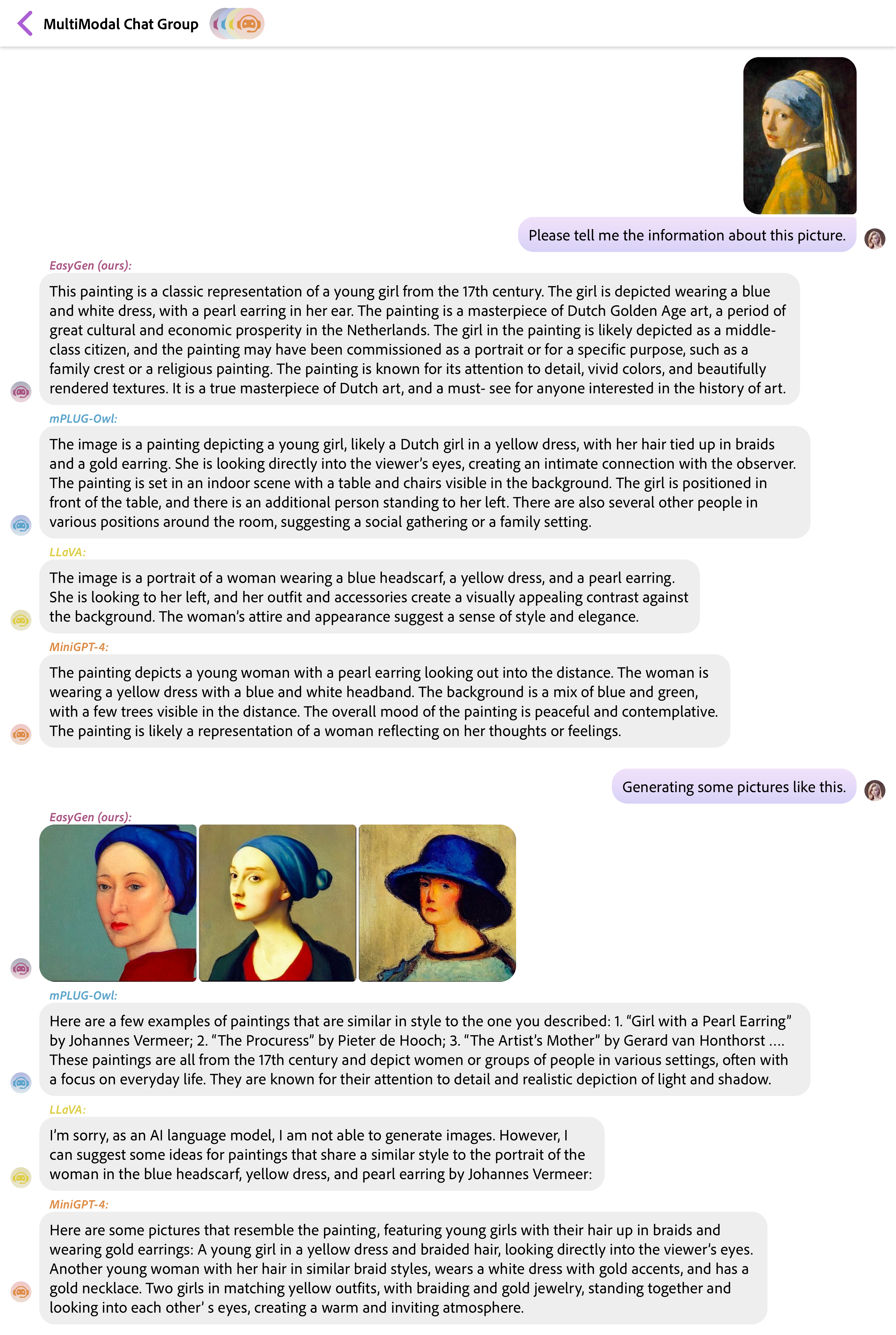}
	\caption{In this case study, for the first question, \modelname{} can give an accurate answer including the background information of the image. With the image generation ability of BiDiffuser, \modelname{} can generate visual responses.}
        \label{llava80K impact image}
\end{figure*}

\begin{figure*}[htbp]
	\centering
	\includegraphics[width=12.5cm, scale=0.8]{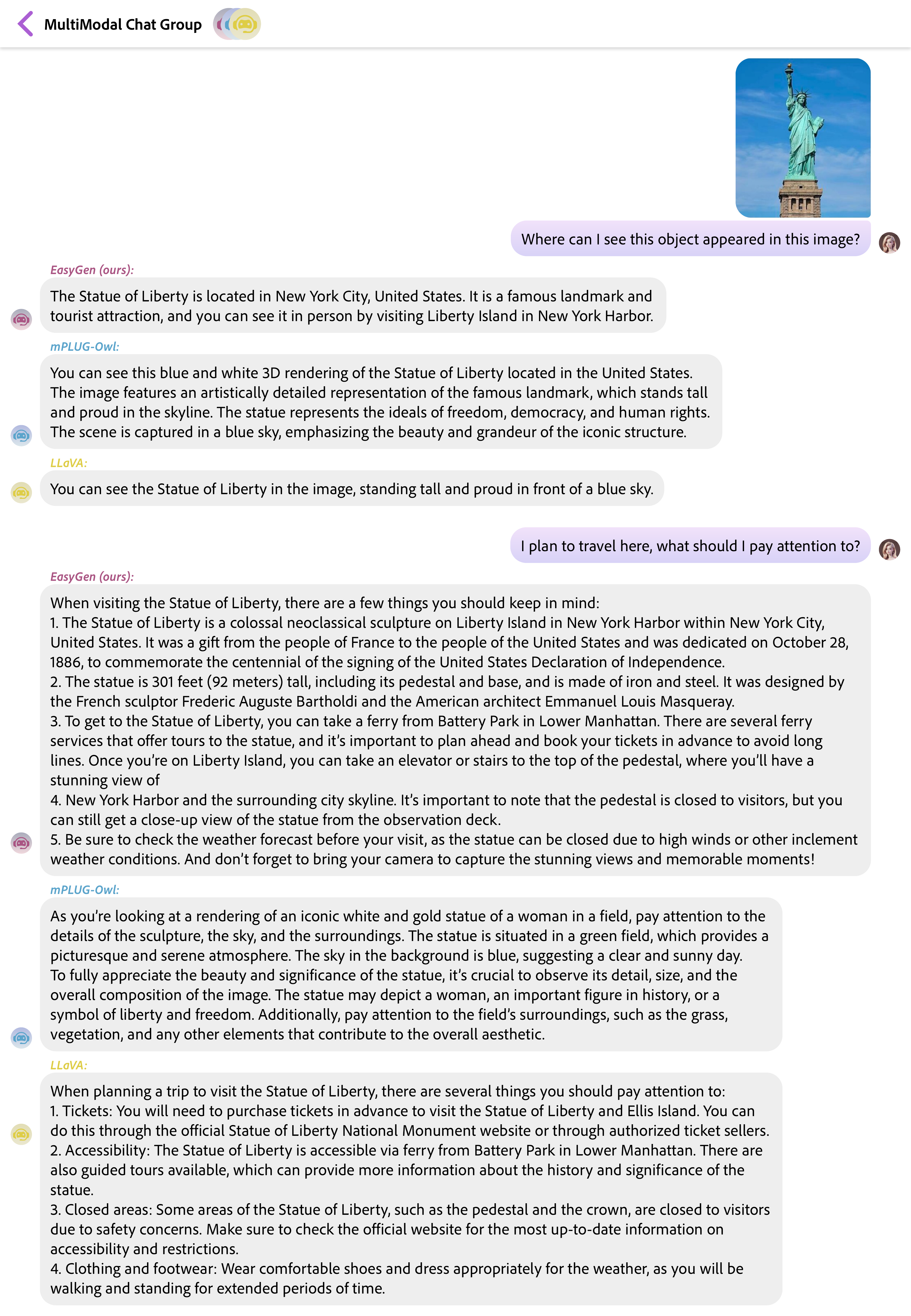}
	\caption{In this case study, for the first question, \modelname{} can give an accurate answer, but the responses of the other two models are a bit biased. For the second question, \modelname{} and LLaVA both give reasonable advice.}
        \label{llava80K impact llava3}
\end{figure*}

\begin{figure*}[htbp]
	\centering
	\includegraphics[width=12.5cm, scale=0.8]{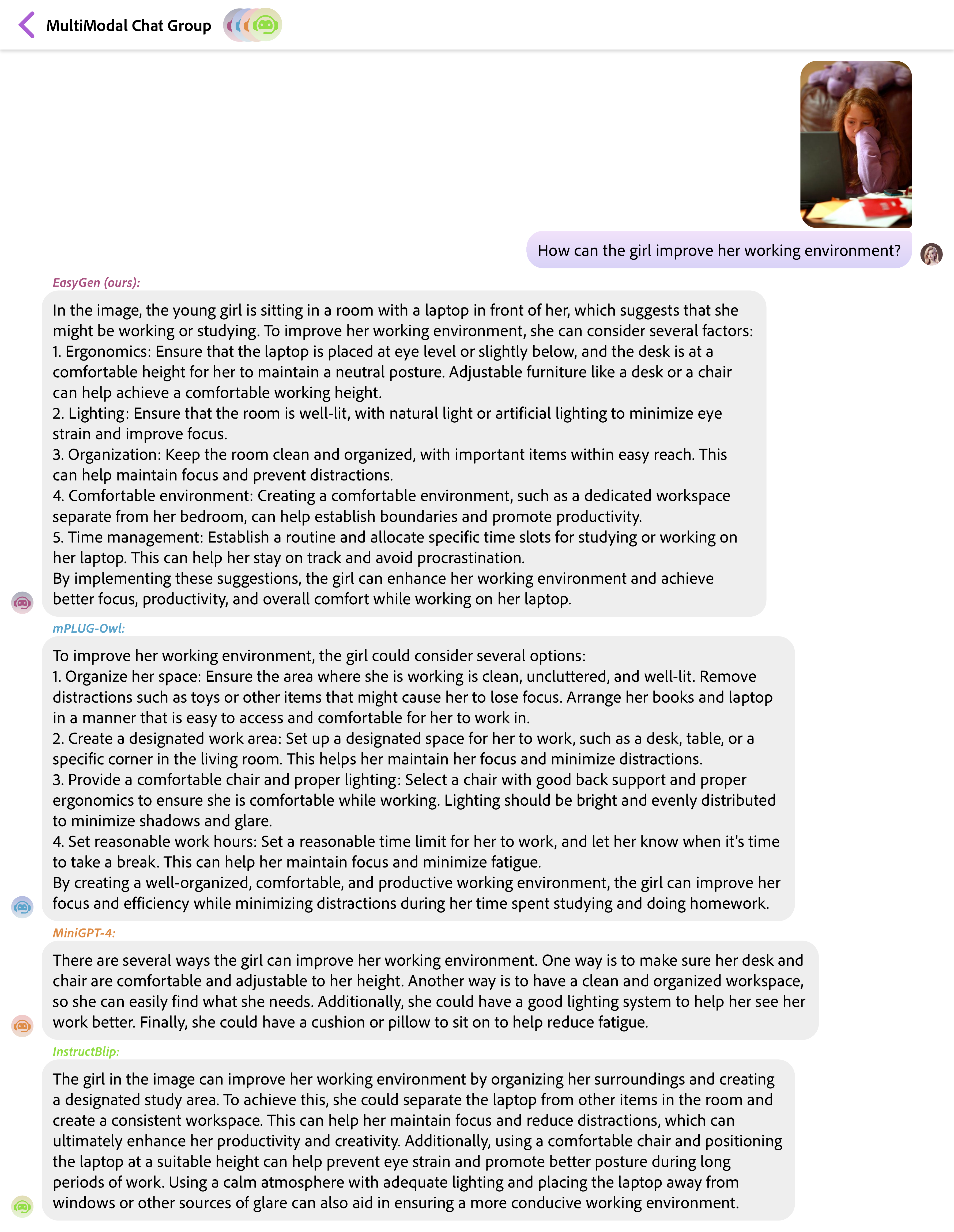}
	\caption{From this example, we can find that the response from \modelname{} is more comprehensive and coherent. This shows \modelname{} can give reasonable suggestions based on the given image.
 }
        \label{llava80K impact llava2}
\end{figure*}

\begin{figure}[b]
	\centering
	\includegraphics[width=7.5cm, scale=0.7]{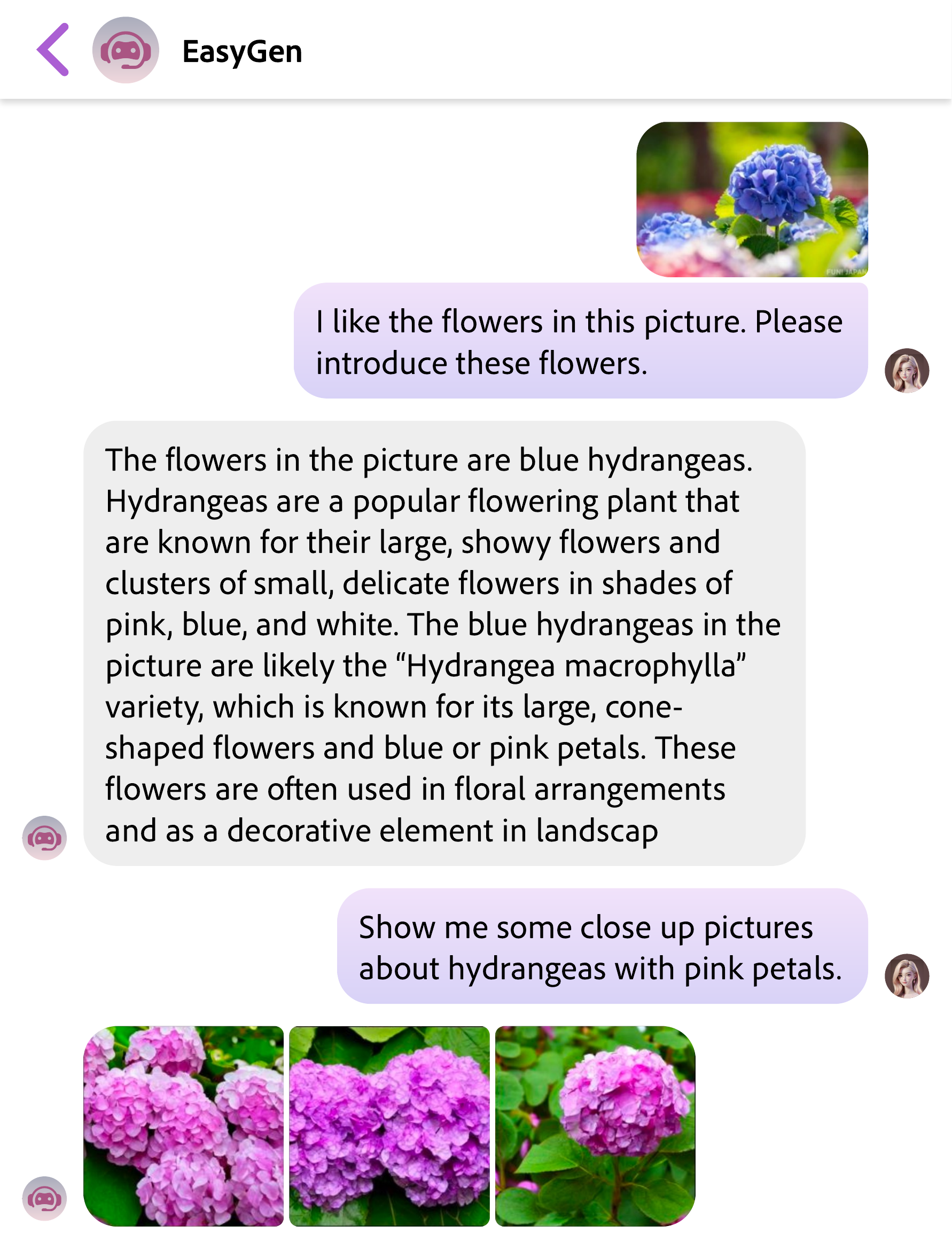}
	\caption{Example of multimodal response generation.}
 \label{easygen_case1}
\end{figure}

\begin{figure}[htbp]
	\centering
	\includegraphics[width=7.5cm, scale=0.8]{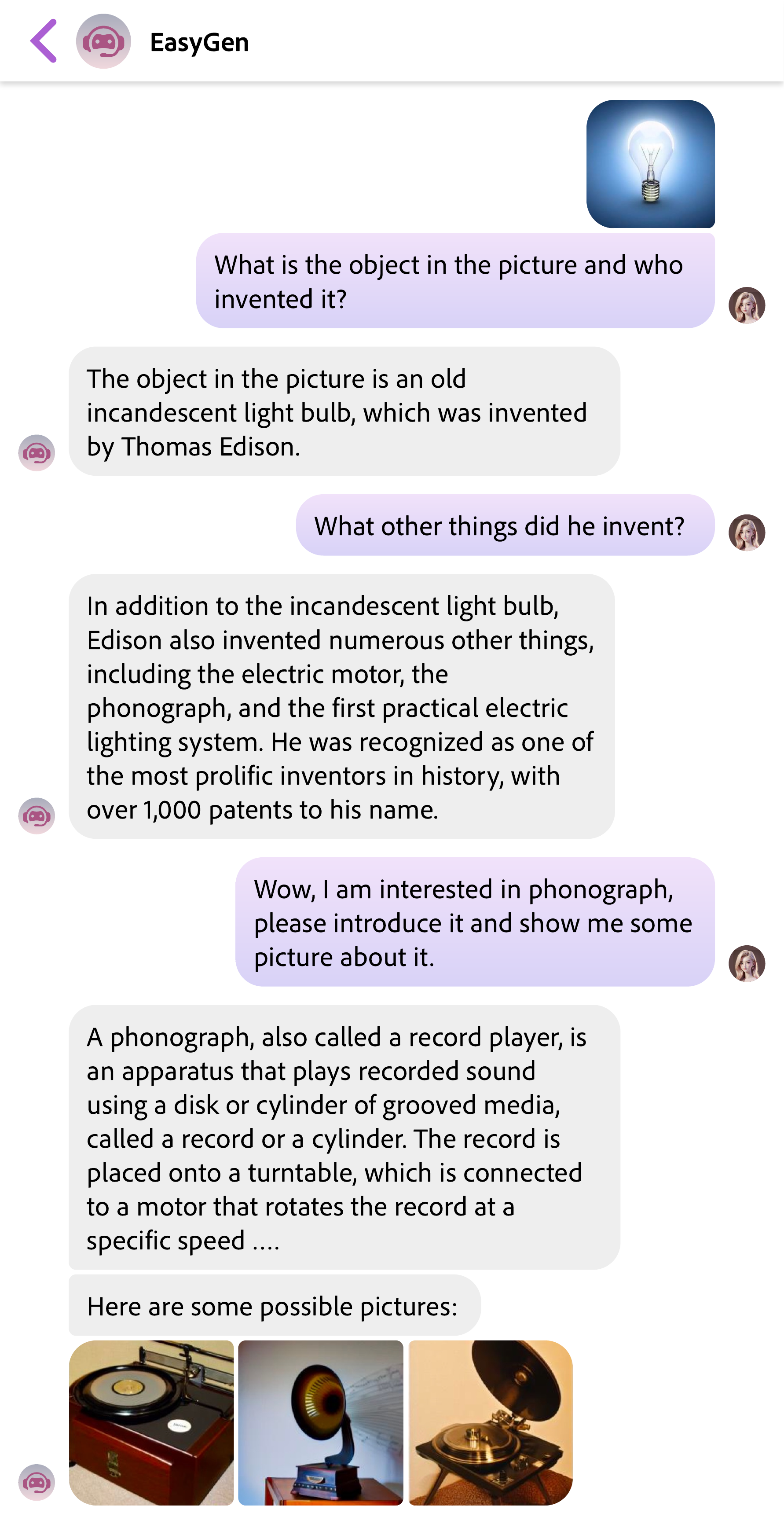}
	\caption{Example of multimodal response generation.}
 \label{easygen_case2}
\end{figure}

We present several instances on PhotoChat dataset in Figure~\ref{tab:case_study} and the image-captioning task in Figure~\ref{example1}.
In Figure~\ref{llava80K impact image}, \ref{llava80K impact llava2}, \ref{llava80K impact llava3}, we compare \modelname{} with state-of-the-art multimodal language models. The responses of MiniGPT-4, LLaVA, mPLUG-owl and InstructBlip are obtained from their official demos. Morever, in Figure~\ref{easygen_case1}, \ref{easygen_case2}, we show \modelname{}'s ability to accept multimodal inputs and generate multimodal responses.


\label{sec:appendix}

\end{document}